\documentclass[runningheads]{llncs}

\usepackage{eccv}

\usepackage{eccvabbrv}

\usepackage{graphicx}
\usepackage{booktabs}

\usepackage[accsupp]{axessibility}  

\usepackage{hyperref}

\usepackage{orcidlink}
\usepackage{url}
\usepackage{algorithm}
\usepackage{algorithmic}
\usepackage{bm}
\usepackage{amsmath}
\usepackage{amssymb}
\usepackage{multicol}
\usepackage{multirow}
\usepackage{pifont}
\usepackage[table,xcdraw]{xcolor}
\usepackage{graphicx}
\usepackage{caption}
\usepackage{subcaption}
\usepackage{wrapfig}
\usepackage{xcolor}
\usepackage{import}

\begin{document}

\title{SAGE: Synchronized Action-Gaze Recognition and Anticipation for Human Behavior Understanding} 

\titlerunning{SAGE}

\author{Chenyi Kuang\inst{1,2}\textsuperscript{*}\orcidlink{0009-0007-2995-2821} \and
Nakul Agarwal\inst{1}\textsuperscript{\dag}\orcidlink{0000-0002-7684-6075} }

\authorrunning{C. Kuang \and N. Agarwal}

\institute{
\textsuperscript{1}\hspace{0.05em}Honda Research Institute, USA \qquad
\textsuperscript{2}\hspace{0.05em}Rensselaer Polytechnic Institute
}

\maketitle
\begingroup
\renewcommand{\thefootnote}{*}
\footnotetext{work done during an internship at Honda Research Institute, USA.}
\renewcommand{\thefootnote}{\dag}
\footnotetext{corresponding author.}
\renewcommand{\thefootnote}{\ddag}
\footnotetext{\url{https://usa.honda-ri.com/exocook}.}
\endgroup

\begin{abstract}
Human object interaction (HOI), gaze pattern, and their anticipation are intricately linked, providing valuable insights into cognitive processes, intentions, and behavior. However, most existing models handle gaze and actions separately, missing both their interdependence and the advantages of a unified solution. This paper presents a novel unified framework, SAGE (\textbf{S}ynchronized \textbf{A}ction-\textbf{G}az\textbf{E}), which integrates simultaneous recognition and anticipation of both HOI and human gaze into a single unified end-to-end trainable model. Our approach leverages a transformer-based architecture and incorporates gaze data into spatiotemporal attention mechanisms to simultaneously predict current and future human actions and gaze behavior. We explore this bidirectional relationship between gaze and actions under different scenarios, whether requiring a close-up, detailed view (egocentric) or a wider, more contextual view (exocentric), making our framework versatile for various applications. Additionally, due to lack of datasets for comprehensive analysis of both HOI and gaze in exocentric videos, we establish a new benchmark \textit{Exo-Cook}\textsuperscript{$\ddag$} 
to facilitate further research in this domain. Experiments on three benchmark datasets—VidHOI, EGTEA Gaze+, and Exo-Cook—show that jointly modeling gaze and actions across current and future frames achieves consistently strong results, often surpassing specialized state-of-the-art models tailored to individual tasks.
By unifying actions and attention in a comprehensive way, our work lays the groundwork for more intuitive human-machine interaction.
\end{abstract}
\section{Introduction}
\label{sec:intro}
\begin{figure}[t!]
    \centering
    \includegraphics[width=0.96\textwidth, height=0.26\textwidth]{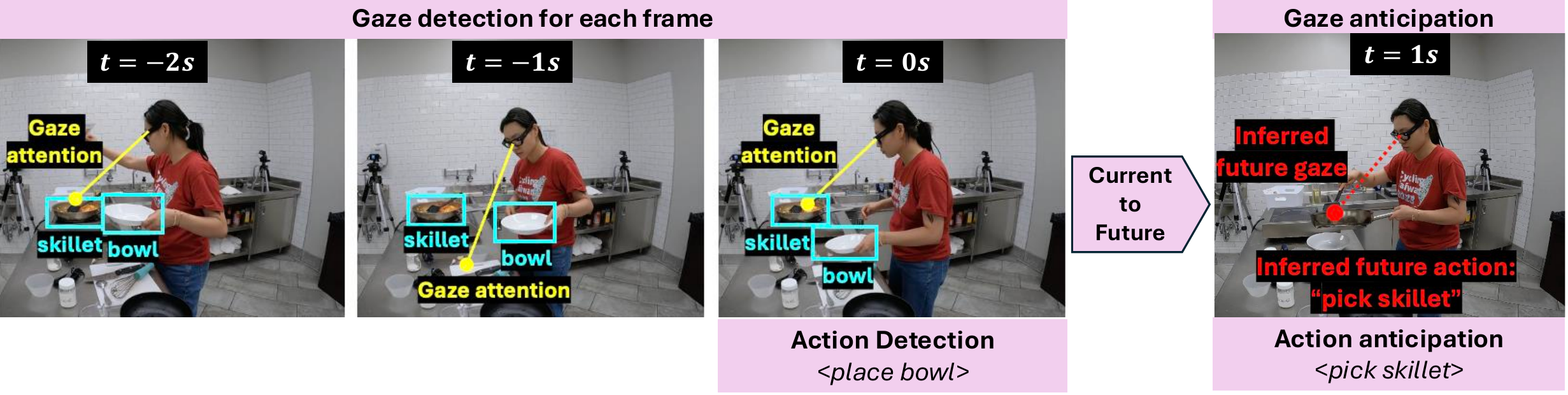}
    \caption{\textbf{Illustration of action-gaze correlation over time.} Gaze reveals both the current focus of attention and future intention, providing complementary cues for recognizing ongoing actions and anticipating subsequent human-object interactions.} 
    \label{fig:intro_figure}
\vspace{-2em}
\end{figure}

Consider a person preparing a meal with a skillet, as shown in Figure~\ref{fig:intro_figure}. Their gaze remains focused on the skillet while they place an empty bowl on the table. Even before any hand movement begins, this visual attention provides a strong cue that the next action is likely to involve picking up the skillet and transferring the food into the bowl. Their gaze is also expected to remain on the skillet, as it is the object involved in the upcoming action. 
Humans perform this reasoning effortlessly. By observing where someone is looking and how they interact with objects, we can simultaneously recognize current actions and anticipate future ones. These abilities arise from a unified cognitive process, highlighting the strong interdependence between gaze and action understanding.
For intelligent systems to interact naturally with people, they must unify perception and prediction—understanding both current and future behavior without relying on separate task-specific pipelines.
From assistive robots to driver monitoring and AR assistants, many applications require understanding current actions and anticipating future ones. Gaze reveals intention, while human–object interactions capture engagement with the environment. Together, they provide rich insight for understanding human behavior.

Yet, most existing approaches treat these tasks separately: some recognize HOI and actions~\cite{ji2021detecting, cong2021spatial, tu2022video, vid-hoi, mascaro2023hoi4abot, ni2023human, hao2022group, wang2020symbiotic} or anticipate them~\cite{InAViT, AVT, AFFT, FHOI, cong2021spatial, ni2023human}, while others model gaze detection~\cite{GLC, li2021eye, huang2018predicting, tafasca2024sharingan, chong2020detecting} or gaze anticipation~\cite{lai2024listen, zhang2017deep}. Few attempt to integrate both, and those that do either use disjointed pipelines or focus only on egocentric views~\cite{li2015delving, ma2016going, singh2016first, ni2023human, EGTEAGaze+, li2021eye, min2021integrating, huang2020mutual}. Importantly, most do not model how gaze and actions evolve together into the future—a key aspect of joint anticipation and intuitive, human-like understanding.
We propose SAGE, a unified end-to-end framework that jointly recognizes and anticipates both human gaze and actions. Built on a transformer backbone, SAGE integrates gaze into spatiotemporal attention to learn shared representations that capture the coupling between where people look and what they do. Central to our design are two modules: the Gaze-Conditioned Spatial Attention (GCSA) module, which injects gaze into spatial attention to highlight human–object interaction cues, and the Gaze-Conditioned Temporal Prediction (GCTP) module, which models temporal correlations between future gaze and actions. SAGE can be trained on either egocentric or exocentric views by adapting only the feature construction stage, while the key components GCSA and GCTP remain consistent.

However, training and evaluating such models demands appropriate datasets. Most existing datasets focus exclusively on either actions~\cite{damen2018scaling, kuehne2014language, lea2016learning} or gaze~\cite{chong2020detecting, tafasca2023childplay}, and the few that include both are primarily limited to first-person perspectives~\cite{li2021eye, grauman2022eg04}. While Ego-Exo4D~\cite{grauman2024ego} provides third-person annotations for both gaze and actions, it lacks a benchmark for joint modeling and evaluation. Moreover, Ego-Exo4D is not directly usable for this task—it requires preprocessing of annotations, new label creation, modality alignment, and task-specific structuring for joint gaze–action modeling, which is non-trivial (see Section~\ref{exo-cook}). To address this, we introduce Exo-Cook, a third-person benchmark derived from Ego-Exo4D, specifically curated for evaluating unified models of gaze-action recognition and anticipation. Exo-Cook fills a critical gap and, to the best of our knowledge, is the first dataset to enable joint gaze-action research in third-person settings.

In summary, our contributions are four fold. \textbf{First}, we propose a unified end-to-end trainable architecture that integrates detection and anticipation of both HOI and gaze, allowing for joint optimization of these tasks for comprehensive human behavior understanding. \textbf{Second}, we introduce a Gaze Conditioned Spatial Attention (GCSA) module that provides HOI cues in the spatial domain and a Gaze Conditioned Temporal Prediction (GCTP) module which simultaneously models temporal correlations between future gaze patterns and future actions. 
\textbf{Third}, unlike prior works that focus on either egocentric or exocentric views, our modular framework supports both through separate training, enabling broader applicability across diverse scenarios.
\textbf{Fourth}, we demonstrate the efficacy of SAGE against state-of-the-art methods and introduce Exo-Cook, a new benchmark for evaluating models that integrate HOI and gaze analysis in third-person videos—providing a comprehensive foundation for future research in this area.
\vspace{-3mm}

\section{Related Works}
\subsection{Gaze Detection and Anticipation}
Gaze detection focuses on estimating a person’s current visual attention point and has been studied in both egocentric and exocentric settings. 
The GazeFollow dataset~\cite{recasens2015they} laid the foundation for static gaze-following in images and was later extended to videos~\cite{recasens2017following,chong2020detecting,tafasca2024sharingan}. Early models typically used two-stream architectures that fused head pose and saliency cues~\cite{chong2018connecting, horanyi2023they}, while recent methods have enhanced gaze localization by incorporating depth information~\cite{tafasca2023childplay, bao2022escnet, GFIE} and human pose~\cite{gupta2022modular}. While the primary goal of these models is accurate gaze localization, gaze is often used downstream to aid action recognition. In contrast, leveraging action to inform gaze has been rarely explored~\cite{huang2020mutual} and may even hurt the performance~\cite{EGTEAGaze+,li2021eye}. In this work, we propose a joint estimation framework to demonstrate that action understanding and gaze detection enhance each other in a complementary manner.

Gaze anticipation focuses on forecasting where gaze will shift in the future. This task has been studied \textit{exclusively} in egocentric settings, where gaze shifts often precede interactions with objects. Existing approaches~\cite{lai2024listen, zhang2017deep} model gaze trajectories over time by leveraging motion cues and temporal scene context. However, these methods treat gaze anticipation as an isolated task and do not account for its interplay with actions. More importantly, to our knowledge, no existing work has explored gaze anticipation in exocentric videos—leaving a critical gap in third-person behavior modeling.

\subsection{Recognizing and Anticipating Actions}
To infer actions from appearance and motion, two-stream models such as I3D-2Stream~\cite{li2021eye} and R34-2Stream~\cite{sudhakaran2018attention} capture RGB and motion cues, while SAP~\cite{wang2020symbiotic}, GC-TSM~\cite{hao2022group}, and ACE~\cite{ghoddoosian2025ace} improve egocentric and procedural understanding. Gaze complements these visual-temporal methods by highlighting task-relevant regions, from early saliency or gaze based approaches with handcrafted features~\cite{wang2016, chen2014, mathe2012, shapovalova2013} to recent HOI models that integrate gaze heatmaps into multimodal transformers~\cite{ni2023human}; related attention-based methods instead learn such regions without explicit gaze supervision~\cite{sharma2016}. 

Meanwhile, action anticipation aims to forecast future behavior from partial observations. 
Existing methods often model HOIs using spatio-temporal graphs~\cite{jain2016structural, materzynska2020somethingelse, ou2022object, teng2021target, wang2018videos}, supported by rich datasets~\cite{AVT, damen2022rescaling, gidar2021raft} and architectural advances like RU-LSTM~\cite{furnari2019}, AVT~\cite{AVT}, MemViT~\cite{fu2022}, RAFTformer~\cite{girase2023latency}, UADT~\cite{guo2024uncertainty}, InAViT~\cite{InAViT} as well as several other methods~\cite{wang2024vamos, mittal2024can, zhao2024antgpt, zhang2024object}. Recent trends also explore goal-conditioned reasoning~\cite{roy2022}, motion primitives~\cite{dessalene2023cherbligs}, and multimodal cues such as audio~\cite{wu2021}. 


\subsection{Joint Action-Gaze Modeling}
Early deep models explored the synergy between gaze estimation and action recognition in first-person video. 
Li et al.~\cite{EGTEAGaze+, li2021eye} proposed one of the first unified frameworks to jointly predict gaze and actions, where predicted gaze maps are used to spatially reweight visual features for action recognition. The model leverages gaze as an attention cue but focuses primarily on recognition in egocentric settings. Huang et al.~\cite{huang2020mutual} introduced a mutual context network that jointly estimates gaze and action by exchanging information between the two tasks. In their gaze-guided action module, predicted gaze positions are used to spatially aggregate visual features for improved action recognition. Both approaches are designed for egocentric settings. In exocentric scenarios, however, gaze must be associated with a specific human in the scene. Prior models \cite{EGTEAGaze+,li2021eye,huang2020mutual} do not explicitly model human location or orientation, making it difficult to determine which subject the predicted gaze corresponds to. Yet, despite its predictive power, gaze remains underutilized in action anticipation. No existing approach models the temporal link between future gaze and future actions, or between current gaze-action modeling with future gaze-action.

\section{Exo-Cook Dataset}
\label{exo-cook}
\begin{figure}[t!]
    \centering
    \includegraphics[width=0.76\textwidth, height=0.30\textwidth]{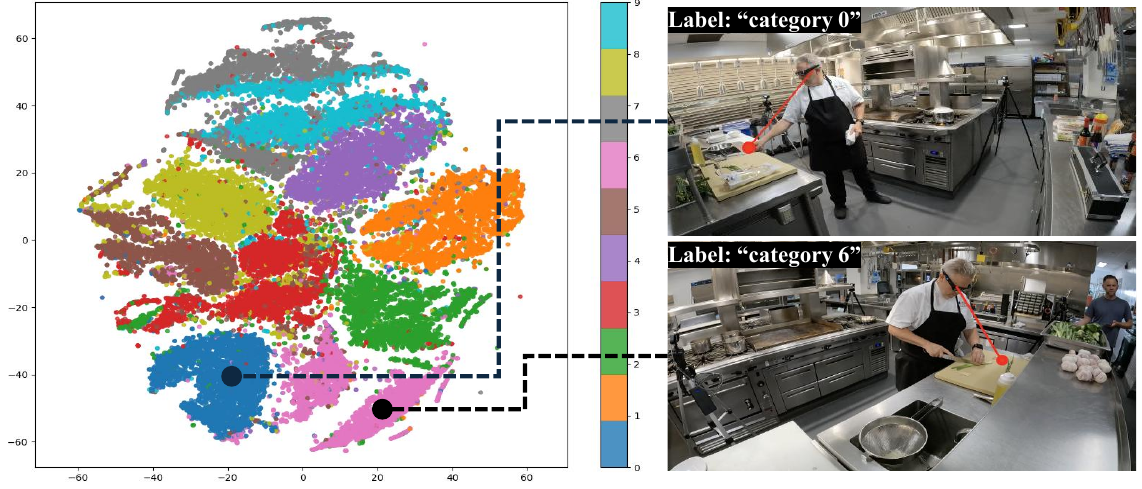}
    \caption{\textbf{Left}: Exo-Cook labels clustered and projected using t-SNE. \textbf{Right}: Visualization examples from two clusters (category 0 and 6) with atomic action descriptions and their verb–noun annotations - (Top) \textit{C picks a strainer from the countertop with his right hand"} [Verb: picks; Noun: strainer]. (Bottom) \textit{“C cuts the green onion with a knife on the cutting board”} [Verb: cuts; Nouns: green onion, knife, cutting board].} 
    \label{fig:clustering-sec3}
\vspace{-1em}
\end{figure}
Very few datasets provide both gaze and action annotations, and those that do are typically limited to egocentric perspectives; moreover, not all scenarios in these datasets are suitable for joint action–gaze analysis. Ego-Exo4D~\cite{grauman2024ego} is a notable exception—a large-scale human-activity dataset with egocentric and multiple (4–5) exocentric views across 8 domains, 740 participants, and 123 scenes—offering time-aligned textual action descriptions and dense annotations (3D pose, gaze, object masks). However, it lacks a benchmark for joint modeling and evaluation. We utilize videos from the \textit{cooking} domain to create our own benchmark, which exhibits strong gaze–action alignment and variability suited for joint action-gaze analysis. In contrast, other seven domains 
show weak correspondence (e.g., soccer, basketball) or diffused gaze patterns (e.g., music), making them less suitable. This choice is consistent with prior works evaluating exclusively on cooking datasets~\cite{li2021eye, huang2020mutual}.
Since the Ego-Exo4D annotations are not directly suitable for model training, we systematically construct Exo-Cook through extensive label generation and task-specific adaptation. Specifically, we extract 658 cooking videos and 189,225 textual descriptions from Ego-Exo4D and curate them through a structured data transformation pipeline as follows:
\textbf{(a) Human bounding boxes:} We use the pipeline in~\cite{ni2023human} with YOLOv5~\cite{jocher2022yolov5} to detect full-body and head boxes of the camera wearer in third-person views.
\textbf{(b) Object bounding boxes:} Boxes are generated for interactable objects from instance masks provided in the Ego-Exo4D metadata.
\textbf{(c) Gaze heatmaps:} We go through four steps to generating gaze heatmaps. 
First, we calculate 3D gaze intersection point $ G^{aria}_{3D}$ from binocular gaze vectors. Second, transform $ G^{aria}_{3D}$ to the third-person camera space via the relative pose, yielding \( G^{camera}_{3D} \). Third, project \( G^{camera}_{3D} \) to 2D as \( G^{camera}_{2D} \). Fourth, generate a 2D Gaussian heatmap \( M^{\text{pseudo}} \), centered at \( G^{camera}_{2D} \), with standard deviation: ${\sigma = \frac{W_{\text{hm}} + H_{\text{hm}}}{2} \cdot \frac{3}{64}}$
where \( (W_{\text{hm}}, H_{\text{hm}}) \) denotes the heatmap size. 
\textbf{(d) Action labels:} We use spaCy~\cite{spacy2024} to extract all candidate verbs and nouns from each textual action description. As illustrated in Figure~\ref{fig:clustering-sec3}, spaCy may return multiple noun options for a single description (e.g., category 6). To identify the most semantically relevant object, we compute the cosine similarity between the BERT~\cite{devlin2019bert} embeddings of each verb–noun pair and that of the full sentence. The verb–noun pair with the highest similarity score is then transformed into a high-dimensional embedding using BERT, which we treat as the action embedding. We apply K-means clustering (\(K=10\)), selected via the Elbow Method, to group semantically similar action embeddings into ten categories (indexed 0–9). The action category labels are manually verified after clustering. We provide semantic consistency analysis of the resulting categories and visualizations of semantic interpretations in section~\ref{supp: section exo-cook label creation}.
%
\textbf{(e) Label alignment:} These labels are aligned with video timestamps, and we use DeepSORT~\cite{wojke2017simple} to track human–object trajectories. Exo-Cook contains 33,321 video clips with atomic action descriptions, from which we remove samples lacking gaze labels, valid head boxes, or proper alignment. This yields 32,050 valid clips annotated with bounding boxes, gaze, and action labels, with 2562, 2741, 1193, 688, 2675, 4167, 4000, 3530, 5500, and 4994 clips for class indices 0–9, respectively. 
\textbf{(f) Viewpoint Diversity:} Exo-Cook includes 8,778, 8,721, 6,429, 6,253, and 1,869 clips from Cam01–Cam05, respectively, following the exocentric multi-camera setup of Ego-Exo4D~\cite{grauman2024ego}. We split the data into 25,650 training, 3,200 validation, and 3,200 test samples. More details are described in section~\ref{supp: section exo-cook}.


\section{SAGE}
\label{sec: SAGE}
\begin{figure*}[t!]
    \centering
    \includegraphics[width=\textwidth, height=0.48\textwidth]{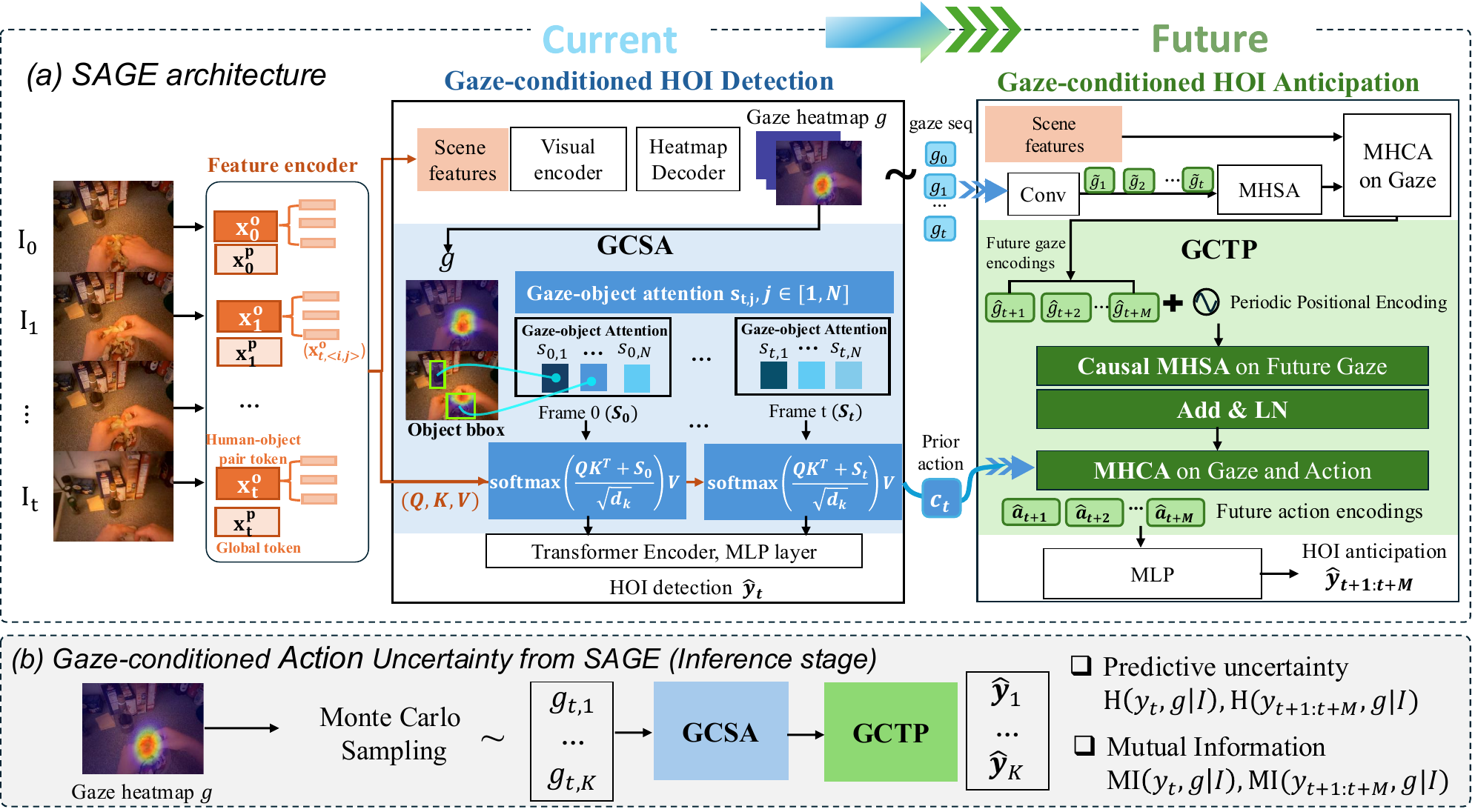}
    \caption{\textbf{Overview of SAGE.} We first present (a) SAGE architecture, which jointly performs gaze detection, HOI detection, and their respective anticipation. A feature encoder extracts human–object interaction tokens, while the gaze detection module predicts a probabilistic heatmap used by \textbf{GCSA} for HOI detection, where \textbf{GCSA} replaces the standard multi-head self-attention (MHSA) operation within the Transformer encoder layer. For forecasting, a Gaze Anticipation module predicts future gaze from observed frames and past gaze, and \textbf{GCTP} models future HOIs conditioned on current interactions and anticipated gaze. We also show (b) gaze-conditioned action uncertainty analysis via sampling from the predicted gaze heatmap at inference.}
    \label{fig:overview}
    \vspace{-2em}
\end{figure*}
In this section, we present the SAGE framework and describe its training objective, as illustrated in Figure~\ref{fig:overview} (a). 
Given a short video clip $I_{0:t}$ where $t$ is the current time step, our objective is to detect the ongoing action at $t$ and anticipate it $M$ time steps into the future. Denoting the action as $y$, the problem can be formulated as computing $p(\bm{y}_{t:t+M} | I_{0:t})$, where $\bm{y}_{t:t+M} = [ y_t, y_{t+1}, \dots, y_{t+M}]$. Note that $y$ in our framework can represent either an HOI triplet or a verb–noun pair, depending on the viewpoint, as both denote \textit{actions}. Egocentric datasets (e.g., EGTEA) adopt the verb–noun pair representation since the human actor is not visible, whereas exocentric datasets (e.g., Vid-HOI) employ HOI triplets due to the visible human presence. We introduce gaze variable $\bm{g}_{0:t+M}$ at corresponding time stamp and reformulate the problem as gaze-conditioned HOI modeling:
{
\setlength{\abovedisplayskip}{3pt}
\setlength{\belowdisplayskip}{3pt}
\begin{equation}
\begin{aligned}
p(\bm{y}_{t:t+M}\mid I_{0:t})
&= \int_G p(\bm{y}_{t:t+M},\bm{g}_{0:t+M}\mid I_{0:t})\,dG \\
&= \int_G p(\bm{y}_{t:t+M}\mid \bm{g}_{0:t+M}, I_{0:t})
\,p(\bm{g}_{0:t+M}\mid I_{0:t})\,dG .
\end{aligned}
\label{eq:marginal likelihood}
\end{equation}
}
%
where $G \triangleq \bm{g}_{0:t+M}$ represents the gaze variable set.
Based on Eq.~\ref{eq:marginal likelihood}, we divide the modeling process into two stages: \emph{present} ($0{:}t$), where video observations are available, and \emph{future} ($t{+}1{:}t{+}M$), where no visual cues are observed. Accordingly, gaze and HOI probabilities are factorized into current and future components.
For gaze, we decompose
$p(\bm{g}_{0:t+M} \mid I_{0:t})
= p(\bm{g}_{0:t} \mid I_{0:t}) \,
p(\bm{g}_{t+1:t+M} \mid \bm{g}_{0:t}, I_{0:t}),$
where the first term models current gaze and the second predicts future gaze trajectories.
For HOI, we factorize $p(\bm{y}_{t:t+M} \mid \bm{g}_{0:t+M}, I_{0:t})
=
p(y_t \mid \bm{g}_{0:t}, I_{0:t}) \,
p(\bm{y}_{t+1:t+M} \mid y_t, \bm{g}_{0:t+M}, I_{0:t}).$
This formulation leads to four probabilistic components in SAGE: gaze detection $p(\bm{g}_{0:t} \mid I_{0:t})$ (SAGE-1), gaze-conditioned HOI detection $p(y_t \mid \bm{g}_{0:t}, I_{0:t})$ (SAGE-2), future gaze prediction $p(\bm{g}_{t+1:t+M} \mid \bm{g}_{0:t}, I_{0:t})$ (SAGE-3), and gaze-conditioned HOI anticipation (SAGE-4) $p(\bm{y}_{t+1:t+M} \mid y_t, \bm{g}_{t+1:t+M}, I_{0:t})$ . 
For model simplification, we assume that future actions $\bm{y}_{t+1:t+M}$ are conditionally independent of past gaze $\bm{g}_{0:t}$, given future gaze $\bm{g}_{t+1:t+M}$ and adopt the approximation: 
$p(\bm{y}_{t+1:t+M} \mid y_t, \ \bm{g}_{0:t+M}, I_{0:t}) \approx
p(\bm{y}_{t+1:t+M} \mid y_t, \bm{g}_{t+1:t+M}, I_{0:t})$.
We validate this causal assumption by evaluating a variant that conditions future action prediction on the full gaze sequence (present and future), which actually performs slightly worse (see section~\ref{supp: sage ablation causal}).
\\
\noindent The integral over $G$ in Eq.~\ref{eq:marginal likelihood} is intractable, so we approximate the integration by expectation over gaze: $p(\bm{y}_{t:t+M}\mid I_{0:t})
\approx \mathbb{E}_{{g_{0:t}, g_{t+1:t+M}}}
\left[p(\bm{y}_{t:t+M}\mid G, I_{0:t}\right])$.
We approximate the expectation $\mathbb{E}$ with $K$ Monte Carlo samples, each sample denoted as $G^{k}=(g_{0:t}^{k}, g_{t+1:t+M}^{k})$.
Finally, the marginal likelihood for current and future HOIs are approximated by sample average $\bar{p}(\bm{y}_{t:t+M}\mid I_{0:t})
=
\frac{1}{K}\sum_{k=1}^K
p(y_t\mid g_{0:t}^{k}, I_{0:t})\;
p(\bm{y}_{t+1:t+M}\mid y_t, \bm{g}_{t+1:t+M}^{k}, I_{0:t})$. 
In practice, MC sampling is enabled during inference to generate multiple gaze perturbations in a single forward pass for action uncertainty estimation and gaze sensitivity analysis (Figure~\ref{fig:overview} (b)). MC Sampling is performed for both current and future gaze; details are provided in section~\ref{main section: uncertainty} and section~\ref{supp: monte carlo sampling}.
\subsection{Feature Encoding and Viewpoint Adaptability} \label{section: feature extraction}
We extract human--object pair tokens from the video using human and object bounding box. HOI tokens are defined as $\mathbf{x}^o_t = [\mathbf{x}_{t,\langle 1,1 \rangle}, \dots, \mathbf{x}_{t,\langle i,j \rangle}, \dots, \mathbf{x}_{t,\langle n_t^h, n_t^o \rangle}],$ where $n_t^h$ and $n_t^o$ denote the numbers of detected humans and objects, and each token encodes the spatial relationship between a human $i$ and an object $j$. To capture global context, we introduce a learnable global token $\mathbf{x}_t^p$ and construct the final input tokens as $\mathbf{X}_{t} =
[\mathbf{x}^o_{t};\mathbf{x}_t^p].$ Note that unlike prior works~\cite{EGTEAGaze+,huang2020mutual,li2021eye}, SAGE can be applied to both egocentric (FPV) and exocentric (TPV) viewpoints using the GCSA mechanism in Eq.~\ref{eq:GCSA} and Eq.~\ref{eq:GCSA feature}. Specifically, for TPV videos, $\mathbf{x}_{t,\langle i,j \rangle}$ incorporates scene features, human body appearance, body location, and human–object spatial relations. Multiple subjects may appear in the scene ($i \geq 1$). In FPV videos, global body location is unavailable, so interaction tokens instead encode relationships between the hands and objects, with the human index fixed to $n_t^h=1$. We show details of constructing the human-object pair token in section~\ref{supp implementation details hoi extraction}. 
\subsection{Modeling Present: Gaze-Conditioned HOI Detection}\label{sec: gaze-conditioned hoi detection}
We first introduce how to model the present (SAGE-1 and SAGE-2) by formulating HOI detection as a \emph{gaze-conditioned} problem. Gaze is represented as a dense spatial probability map and explicitly modulates spatial attention for predicting HOIs. Given the observed frames $I_{0:t}$, the gaze detection module predicts a normalized gaze heatmap $g_t \in \mathbb{R}^{64\times64}, \qquad \sum_{h,w} g_t(h,w) = 1 ,$ which represents a probabilistic distribution over gaze locations at time $t$. We adapt existing gaze detection models \cite{GLC,chong2020detecting,tafasca2024sharingan,ryan2025gazelle} into SAGE architecture for predicting gaze heatmaps.
\\
\textbf{Gaze-Conditioned Spatial Attention (GCSA).} 
Gaze is known as an important human attention cue for analyzing human action and intent. Prior joint gaze–action models~\cite{EGTEAGaze+,huang2020mutual, li2021eye} typically use gaze heatmaps to spatially reweight visual features at pixel level for action recognition. In contrast, we introduce GCSA, which integrates gaze into transformer attention through \textbf{object-level} gaze relations. For each object $j$ in frame $t$ with bounding box, SAGE assigns a gaze relevance score $s_{t,j}$ computed as $s_{t,j}=\mathrm{Norm} \ \{\sum_{(h,w)\in \mathrm{bbox}_{t,j}} g_t(h,w)\}.$
which measures the probability mass of gaze falling inside the object region. Collecting all scores yields a gaze relevance vector $\mathbf{s}_t$, which is formulated into an additive attention bias matrix $\mathbf{S}_t \in \mathbb{R}^{(N+1)\times (N+1)}$, where $N$ is the maximum number of human--object pair tokens.
We inject the gaze bias $\mathbf{S}_t$ into the self-attention operation to form Gaze-Conditioned Spatial Attention:
\begin{equation}
\mathrm{GCSA}(Q_t,K_t,V_t,\mathbf{S}_t)
=
\mathrm{Softmax}\!\left(
\frac{Q_tK_t^\top + \mathbf{S}_t}{\sqrt{d_k}}
\right)V_t .
\label{eq:GCSA}
\end{equation}
By augmenting the attention logits rather than directly masking features \cite{EGTEAGaze+,huang2020mutual,li2021eye}, GCSA biases the model towards interactions involving gaze-relevant objects while still preserving contextual cues from the full scene. 
GCSA replaces the standard multi-head self-attention (MHSA) operation within each Transformer encoder layer. Specifically, the gaze-conditioned attention bias is added to the attention logits before the softmax operation, allowing gaze--object relations to guide self-attention while preserving the standard Transformer architecture.
Our model learns gaze-aware HOI representations by stacking multiple transformer encoder layers (layer number $N_d$) equipped with GCSA. The $\text{TransformerEnc}_{N_d}$ takes the initial human–object tokens as input, equipped with the GCSA module in the MHSA layer. The GCSA-equipped Transformer Encoder and the resulting gaze-conditioned HOI feature $\bm{c}_t$ are denoted as
\begin{equation}
    \bm{c}_t = \text{TransformerEnc}_{N_d} (\mathrm{GCSA}(Q_t,K_t,V_t,\mathbf{S}_t), \mathbf{X}_t)
    \label{eq:GCSA feature}
\end{equation}
\noindent\textbf{HOI Detection.}
We use one simple transformer encoder layer to model the temporal feature in the gaze-aware features $\{\mathbf{c}_0,\dots,\mathbf{c}_t\}$. Since gaze biases the spatial interactions at each frame, the temporal encoder propagates gaze-informed cues over time. Following by a MLP classifier $\phi(\cdot)$, the action probability is predicted as $p(y_t \mid \bm{g}_{0:t}, I_{0:t})
=
\mathrm{Softmax}\!\left(\phi(\mathrm{TransformerEnc}_{N_d}([\mathbf{c}_0,\cdots,\mathbf{c}_t]))\right)$. \vspace{-3mm}

\subsection{Modeling Future: Gaze-conditioned HOI Anticipation}
Prior joint gaze–action models such as~\cite{li2021eye,huang2020mutual} focus on \emph{only recognition}, that only model the correlation between gaze and HOI at the same time step but do not consider how gaze and HOI jointly evolve over time. 
In contrast, SAGE models the temporal relation between future gaze and HOI given current action.
Under the factorization in Eq.~\ref{eq:marginal likelihood}, HOI anticipation is modeled as $
p(\bm{y}_{t+1:t+M} \mid I_{0:t})
=
\int
p(\bm{y}_{t+1:t+M} \mid y_t, \bm{g}_{t+1:t+M}, I_{0:t})
\, p(\bm{g}_{t+1:t+M} \mid \bm{g}_{0:t}, I_{0:t})
\, dG$
which are decomposed into two steps.\\
\textbf{Future Gaze Modeling.} 
Our first estimate the probability
$p(\bm{g}_{t+1:t+M} \mid \bm{g}_{0:t}, I_{0:t})$, where we use historical gaze observations and video appearance feature to model future gaze. 
Each gaze heatmap $g_k \in \mathbb{R}^{64\times64}$ is embedded into a latent token representation
added by periodic positional encoding (PPE) through $\tilde{\mathbf{g}}_k = \mathrm{Conv}(g_k) + \mathrm{PPE}(k), \quad k=0,\dots,t$.
To model gaze dynamics, we apply a Transformer Encoder to capture temporal dependencies in gaze trajectories: $\mathbf{h}^g_{0:t} = \mathrm{MHSA}(\tilde{\mathbf{g}}_{0:t})$. 
Importantly, future gaze is influenced by the HOI feature in the video context. We incorporate historical gaze feature and human-object feature through cross-attention layer, followed by a downstream decoder to generate future gaze heatmaps
$\hat{g}_{t+1:t+M} = \mathcal{D}_\phi(\text{MHCA(}\mathbf{h}^g_{0:t}, {\mathbf{X}}^o_{0:t}))$.
\\
\textbf{Gaze-Conditioned Temporal Prediction (GCTP).}
We propose GCTP to parameterize the conditional distribution
$p(\bm{y}_{t+1:t+M} \mid y_t, \bm{g}_{t+1:t+M}, I_{0:t})$
by conditioning interaction features on predicted future gaze representations. Given the gaze-conditioned interaction feature ${\mathbf{c}}_t$ produced by the GCSA module and predicted future gaze heatmaps, we encode the gaze sequence using a temporal attention module to capture correlations among future gaze shifts. \\
GCTP aligns future gaze representations with the current interaction features through cross-attention, allowing predicted gaze trajectories to guide the anticipation of future HOIs. This operation aligns future gaze dynamics with the interaction-aware scene representation, enabling the model to anticipate actions that are consistent with predicted attention shifts. The gaze-conditioned future HOI encodings are represented as:
\begin{equation}
    \hat{\mathbf{a}}_{t+1:t+M} = \mathrm{TransformerEnc}_{N_d}(\mathrm{MHCA}(\mathrm{MHSA}(\hat{\mathbf{g}}_{t+1:t+M}),\mathbf{c}_t).
\end{equation}
The final future HOI at $t+M$ are produced by a MLP classifier $\hat{y}_{t+M} = \mathrm{Softmax}(\mathrm{MLP}(\mathbf{\hat{a}}_{t+M}))$.
By explicitly modeling future gaze as a conditioning variable, GCTP bridges current gaze–action recognition with future action anticipation. Similar to GCSA, GCTP mechanism can be applied to both egocentric and exocentric videos. To the best of our knowledge, this is the first framework that jointly models current and future gaze–action dependencies for HOI anticipation.

\subsection{Loss function}
    \noindent\textbf{Heatmap Loss} \( \mathcal{L}_{\text{hm}} .\) As we are training a joint model for multiple tasks, we use a simple yet effective pixel-wise MSE (L2) loss for gaze heatmap~\cite{tafasca2024sharingan, tafasca2023childplay} to ensure stable gradients between the predicted heatmap \( g \) and ground truth heatmap \( g^{\text{gt}} \). Two separate MSE loss terms are applied to gaze detection and anticipation module, respectively: $\mathcal{L}_{\text{hm},1} = \sum_{m=0}^{t} \| \hat{g}_m - g^{\text{gt}}_m \|_2^2$, $\mathcal{L}_{\text{hm},2} = \sum_{m=t+1}^{t+M} \| \hat{g}_m - g^{\text{gt}}_m \|_2^2$.
    \noindent\textbf{Action Loss} \( \mathcal{L}_{\text{act}} .\) We apply Cross-Entropy loss for HOI detection $\mathcal{L}_{\text{act},1} = - y^{gt}_m \log \hat{y}_m, \quad$, and it is similarly defined and summed over intermediate frames for HOI anticipation $\mathcal{L}_{\text{act},2}= - \sum_{m=t+1}^{t+M} y^{gt}_m \log \hat{y}_m$.\\
    \noindent\textbf{In-Out Loss} \( \mathcal{L}_{\text{io}} .\) For exocentric cases, SAGE predicts a variable $o \in \mathbb{R}^1$ along with the gaze heatmap at each future step, representing the probability of gaze located within the frame in the future. The in-out loss is defined as the binary Cross-Entropy between the predicted $\hat{o}$ and the ground truth label \( o^{\text{gt}} \) for future gaze $\mathcal{L}_{\text{io}} = \sum_{m=t+1}^{t+M} - o^{\text{gt}}_m \log(\hat{o}_m) - (1 - o^{\text{gt}}_m) \log(1 - \hat{o}_m)$. \\ 
The total loss function for training SAGE model formulates as $\mathcal{L} = \lambda_1 \mathcal{L}_{hm,1} + \lambda_2 \mathcal{L}_{hm,2} + \lambda_3 \mathcal{L}_{io} + \lambda_4 \mathcal{L}_{act,1} + \lambda_5 \mathcal{L}_{act,2}$.

\section{Experiments}

\noindent\textbf{Datasets \& Metrics.} 
To evaluate our framework, we use three datasets. \textbf{Vid-HOI}~\cite{vid-hoi}, an exocentric dataset for video-based HOI detection and anticipation, provides annotated sequences of human–object interactions. \textbf{EGTEA Gaze+}~\cite{EGTEAGaze+}, widely used for egocentric gaze and action tasks, contains over 28 hours of video across 86 action classes from 32 participants. \textbf{Exo-Cook}, introduced in this work, is the first exocentric dataset curated for joint HOI and gaze analysis. Following~\cite{ni2023human}, we evaluate HOI detection and anticipation on Vid-HOI and Exo-Cook using mean average precision (mAP), top-5 recall, precision, accuracy, and F1-score. For egocentric action recognition, we report Top-1 Accuracy on split 3 and the average on all three splits \cite{hao2022group} for fair comparison. For action anticipation, we report mean class accuracy~\cite{InAViT}, while gaze detection and anticipation are evaluated with F1-score, recall, and precision.

Training proceeds in two stages: we first train SAGE-12 for 5 epochs using gaze and action annotations from the current sequence, then initialize SAGE with these weights and continue training using future gaze and action labels. The complete model is trained for 25 epochs in total. Although efficiency is not our focus, SAGE achieves strong recognition and anticipation accuracy while maintaining high computational efficiency, processing a 20-frame input sequence in just 0.63s on an NVIDIA RTX A6000. Further implementation details and efficiency comparisons are provided in section~\ref{supp implementation details} and section~\ref{efficiency}, respectively.

\begin{table}[t!]
\small
\centering
\setlength{\tabcolsep}{1pt}
\caption{\textbf{Ablation study of GCSA and GCTP} under task-based supervision on EGTEA Gaze+. "w/o" and "w/" denote without and with the module, respectively.}
\scalebox{0.74}{
\begin{tabular}{c c|ccc|c|ccc|c}
\toprule
\multirow{2}{*}{} & 
\multirow{2}{*}{Models} & 
\multicolumn{3}{c|}{\begin{tabular}[c]{@{}c@{}}Gaze \\ Detection\end{tabular}} & 
\begin{tabular}[c]{@{}c@{}}Action \\ Recognition\end{tabular} & 
\multicolumn{3}{c|}{\begin{tabular}[c]{@{}c@{}}Gaze \\ Anticipation\end{tabular}} & 
\begin{tabular}[c]{@{}c@{}}Action \\ Anticipation\end{tabular} \\
\cline{3-10}
& & F1 & Rec & Prec & Acc & F1 & Rec & Prec & Acc \\
\hline

\multirow{4}{*}{\rotatebox{90}{Current}} 
& SAGE-1                  & 44.8 & 61.2 & 35.3 & -    & -    & -    & -    & -    \\
& SAGE-2                  & -    & -    & -    & 63.1 & -    & -    & -    & -    \\
& SAGE-12 (w/o GCSA)  & 44.8 & 61.4 & 35.3 & 63.0 
& - & - & - & - \\
& SAGE-12 (w/ GCSA)         & 46.3 & 61.9 & 36.4 & 63.9 &      &      &      &      \\
\hline

\multirow{4}{*}{\rotatebox{90}{Future}} 
& SAGE-3                  & -    & -    & -    & -    & 35.4 & 49.5 & 28.2 & -    \\
& SAGE-4                  & -    & -    & -    & -    &  -   &  -   &  -   & 47.3 \\
& SAGE-34 (w/o GCTP)  & - & - & - & - 
& 34.8 & 49.4 & 27.4 & 44.6 \\
& SAGE-34 (w/ GCTP)         & -    & -    & -    & -    & 35.2 & 49.4 & 27.5 & 52.6 \\
\hline
& SAGE (w/o GCSA, GCTP) 
& 44.8 & 61.4 & 35.3 & 63.1 
& 34.5 & 48.8 & 26.7 & 44.0 \\
& SAGE (w/ GCSA, GCTP) 
& \textbf{46.8} & \textbf{62.1} & \textbf{36.8} & \textbf{65.0} 
& \textbf{37.0} & \textbf{54.9} & \textbf{32.7} & \textbf{58.4} \\
\bottomrule
\end{tabular}
}
\label{table:ablation}
\vspace{-1em}
\end{table}

\subsection{Ablation Study}
We conduct an ablation study on EGTEA Gaze+ in Table~\ref{table:ablation} to disentangle the effects of joint supervision and the proposed GCSA and GCTP modules. Besides the original SAGE components SAGE-1--4, SAGE-12, SAGE-34 and the full SAGE model, we further construct joint-training baselines by removing GCSA and GCTP while keeping identical multi-task supervision. This isolates whether the gains arise from the proposed gaze-conditioned mechanisms rather than joint optimization alone.
For current gaze--action modeling, SAGE-12 without GCSA jointly optimizes gaze detection and action recognition but does not explicitly model their dependency. Introducing GCSA improves both gaze detection (F1: 44.8$\rightarrow$46.3) and action recognition (Acc: 63.0$\rightarrow$63.9), demonstrating the benefit of gaze-conditioned spatial reasoning beyond joint supervision.
Similarly, SAGE-34 without GCTP jointly learns future gaze and action independently. Incorporating GCTP substantially improves action anticipation (Acc: 44.6$\rightarrow$52.6) while slightly improving gaze anticipation, showing that conditioning future actions on predicted future gaze provides informative temporal cues.
Finally, we compare the full SAGE model with a joint-training baseline that removes both GCSA and GCTP while retaining all four task losses. Despite identical supervision, the proposed modules consistently improve gaze detection (F1: 44.8$\rightarrow$46.8), action recognition (Acc: 63.1$\rightarrow$65.0), gaze anticipation (F1: 34.5$\rightarrow$37.0), and action anticipation (Acc: 44.0$\rightarrow$58.4). These results demonstrate that the improvements are not merely due to multi-task training, but stem from explicitly modeling gaze--action dependencies through GCSA and GCTP.
\subsection{SAGE on Exocentric Videos} \label{section:exp-hoi}
\begin{figure}[t]
    \centering
    \begin{minipage}{0.49\linewidth}
        \centering
        \includegraphics[width=\linewidth]{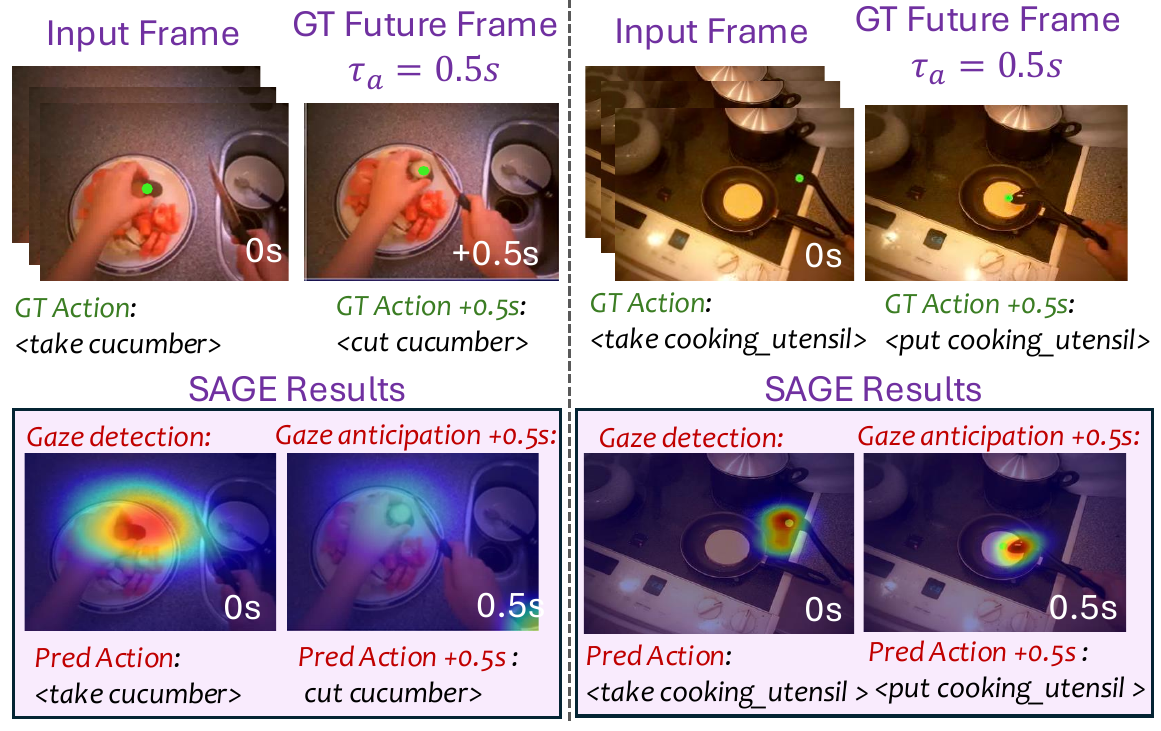}
    \end{minipage}
    \hfill
    \begin{minipage}{0.49\linewidth}
        \centering
        \includegraphics[width=\linewidth]{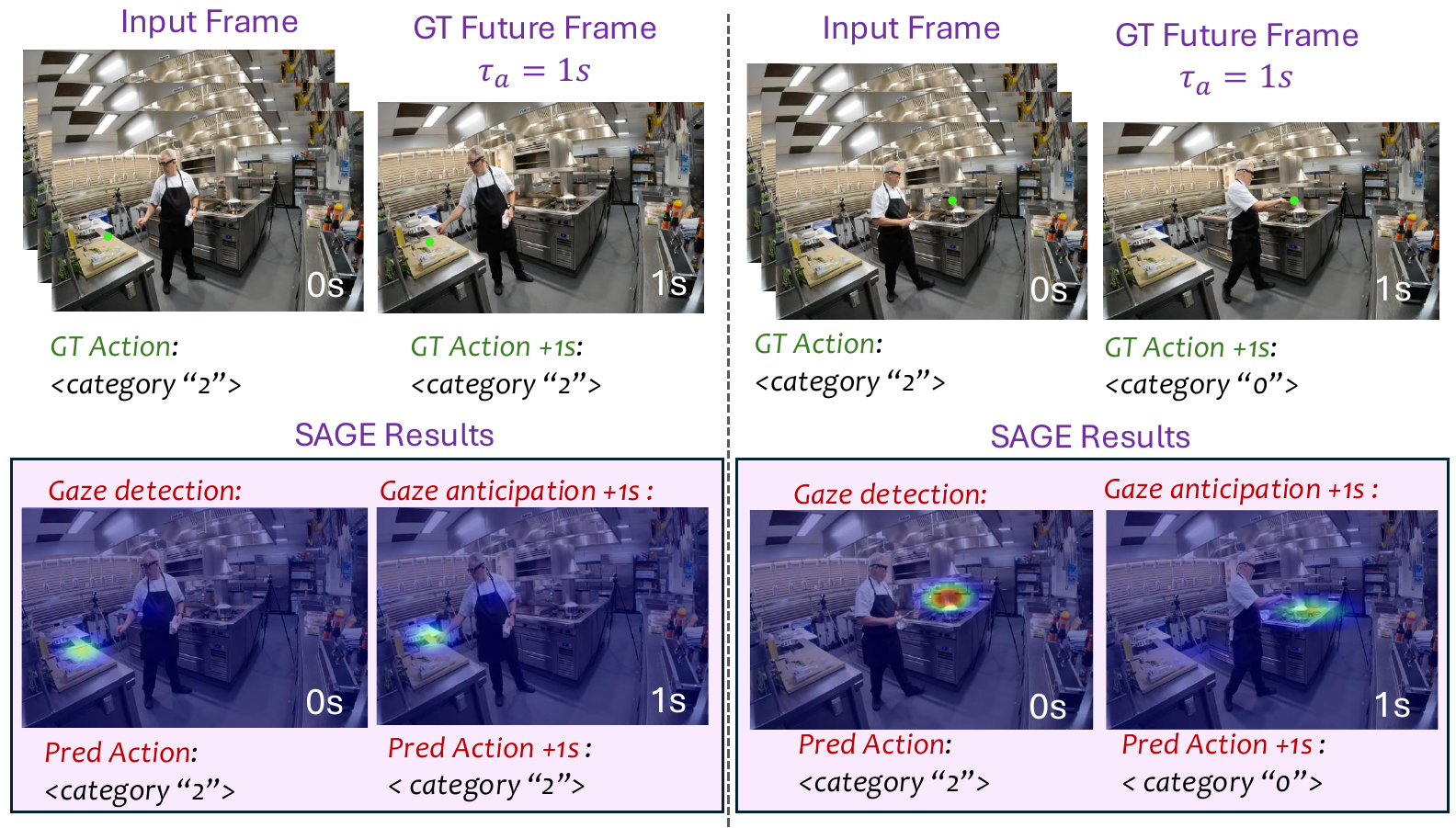}
    \end{minipage}
    \caption{\textbf{Qualitative results.} SAGE on EGTEA Gaze$+$ (left) and Exo-Cook (right), including action and gaze predictions for current and future frames for two samples each. The green dot in first rwo represents ground truth gaze.}
    \label{fig:vis_combined}
    \vspace{-2em}
\end{figure}

\begin{table*}[t]
\small
\centering
\setlength{\tabcolsep}{1.5pt}
\caption{\textbf{SAGE results on Exo-Cook dataset.} $^\dagger$adapted for gaze anticipation. }
\scalebox{0.75}{
\begin{tabular}{c|c|ccc|ccccc|ccc|ccccc}
\toprule
\multirow{2}{*}{Method}                       & \multirow{2}{*}{$\tau_a$} & \multicolumn{3}{c|}{\begin{tabular}[c]{@{}c@{}}Gaze \\ Detection\end{tabular}} & \multicolumn{5}{c|}{HOI Detection} & \multicolumn{3}{c|}{\begin{tabular}[c]{@{}c@{}}Gaze \\ Anticipation\end{tabular}} & \multicolumn{5}{c}{HOI Anticipation} \\ \cline{3-18} 
                                              &                           & F1                        & Rec                      & Prec                     & mAP   & Rec   & Prec & Acc  & F1   & F1                        & Rec                       & Prec                      & mAP   & Rec   & Prec  & Acc   & F1   \\ \hline
VideoAttn~\cite{chong2020detecting}           & -                         & 55.6                      & 72.8                     & 54.8                     & -     & -     & -    & -    & -    & -                         & -                         & -                         & -     & -     & -     & -     & -    \\
Sharingan~\cite{tafasca2024sharingan}         & -                         & 56.6                      & 72.8                     & 55.3                     & -     & -     & -    & -    & -    & -                         & -                         & -                         & -     & -     & -     & -     & -    \\
Gaze-LLE \cite{ryan2025gazelle} & - & 56.8 & 72.4 & 55.0 & - & - & - & - & - & - & - & - & - & - & - & - & - \\
ST-Gaze~\cite{ni2023human}                    & -                         & -                         & -                        & -                        & 42.5  & 77.0  & 61.9 & 59.0 & 66.9 & -                         & -                         & -                         & -     & -     & -     & -     & -    \\
VideoAttn~\cite{chong2020detecting}$^\dagger$   & -                         & -                         & -                        & -                        & -     & -     & -    & -    & -    & 42.2                      & 54.6                      & 42.5                      & -     & -     & -     & -     & -    \\
Sharingan~\cite{tafasca2024sharingan}$^\dagger$ & -                         & -                         & -                        & -                        & -     & -     & -    & -    & -    & 40.1                      & 52.7                      & 40.4                      & -     & -     & -     & -     & -    \\
Gaze-LLE \cite{ryan2025gazelle}$^\dagger$ & - & - & - & - & - & - & - & - & - & 38.6 & 50.3 & 36.2 & - & - & - & - & - \\
ST-Gaze~\cite{ni2023human}                    & 1                         & -                         & -                        & -                        & -     & -     & -    & -    & -    & -                         & -                         & -                         & 41.1  & 76.2  & 58.2  & 53.7  & 66.1 \\ \hline
SAGE (VideoAttn)                              & 1                         & 55.8                      & 73.2                     & \textbf{55.7}                     & 44.2  & 79.0  & 62.6 & 58.5 & 68.9 & 48.5                      & 62.1                      & 49.6                      & 42.4  & 76.6  & 59.0  & 54.6  & 66.9 \\
SAGE (Sharingan)                              & 1                         & \textbf{57.7}                      & \textbf{73.4}                     & 55.5           &    \textbf{44.6}      &   \textbf{79.4}    &   \textbf{63.1}    &      \textbf{60.2} &   \textbf{69.4}   & \textbf{49.2}                      & \textbf{62.8}                      & \textbf{50.0}                      &    \textbf{42.5}   &    
\textbf{76.8}   &    \textbf{59.2}   & \textbf{54.8}  &   \textbf{67.2}  \\
\bottomrule
\end{tabular}}
\vspace{-1.5em}
\label{table: SAGE on Exo-Cook}
\end{table*}
\noindent\textbf{Exo-Cook Dataset.} We evaluate SAGE on four tasks, as summarized in Table~\ref{table: SAGE on Exo-Cook}. 
Since no existing methods directly benchmark on this dataset, we create multiple baseline models by adapting existing models on Exo-Cook.  
We establish three baselines for gaze detection and anticipation, by training \& evaluating three state-of-the-art gaze detection models VideoAttn~\cite{chong2020detecting}, Sharingan~\cite{tafasca2024sharingan} and Gaze-LLE~\cite{ryan2025gazelle} on Exo-Cook. Among them, 
VideoAttn~\cite{chong2020detecting} and Sharingan~\cite{tafasca2024sharingan} are incoporated into SAGE architecture, noted as SAGE (VideoAttn) and SAGE (Sharingan). 
We run ST-Gaze model~\cite{ni2023human} on Exo-Cook to establish the first baseline for HOI detection and anticipation. For anticipation setting, we set the anticipation horizon $\tau_a=1s$ and present results for additional horizons in section~\ref{supp anticipation horizon}.
SAGE (Sharingan) outperforms ST-Gaze on HOI detection by +2.1 mAP and +1.0 accuracy and achieves highest F1 score (69.4). Similarly, on HOI anticipation, SAGE (Sharingan) has the best overall performance, with an F1 score of 67.2 and mAP of 42.5—surpassing the ST-Gaze baseline (F1: 66.1, mAP: 41.1). 
In the gaze detection task, SAGE (Sharingan) achieves the highest F1 score (57.7), slightly outperforming the best baseline Sharingan (56.6) by +1.1 points. While precision is similar (55.5 vs. 55.3), SAGE shows a modest gain in recall (+0.6). For gaze anticipation, SAGE (Sharingan) achieves highest F1 score of 49.2, recall of 62.8, and precision of 50.0, outperforming the closest baseline (ST-Gaze). By comparing SAGE (VideoAttn) and SAGE (Sharingan), we show that integrating more advanced gaze model architecture helps improve the joint model performances. We show some qualitative results in Figure~\ref{fig:vis_combined}, and provide more detailed qualitative results in section~\ref{supp: section vis}.\\
\noindent\textbf{Vid-HOI Dataset.} We take ST-Gaze~\cite{ni2023human} and STTran~\cite{cong2021spatial} as baseline models for HOI detection and anticipation in Table~\ref{table:HOI Vid detection} and Table~\ref{table:HOI Vid anticipation}. Gaze is \textit{not} evaluated on Vid-HOI as no gaze labels are provided. 
For HOI detection, SAGE achieves the best mAP, precision and F1 values. The improved performance of ST-Gaze$^*$~\cite{ni2023human} in recall and Accuracy is attributed to the use of word embeddings as an additional input modality. We individually analyze the impact of GCSA in section~\ref{supp: section experiments}. For HOI anticipation, SAGE outperforms STTran~\cite{cong2021spatial} and ST-Gaze~\cite{ni2023human} across all anticipation horizons (1s, 3s, 5s) for mAP, precision, Accuracy and F1. 

\begin{table*}[t]
\small
\centering

\begin{minipage}[t]{0.48\textwidth}
    \centering
    \setlength{\tabcolsep}{1pt}
    \caption{\textbf{HOI detection} in Oracle mode on Vid-HOI dataset. * includes word embedding module.}
    \scalebox{0.8}{
    \begin{tabular}{c c c c c c c}
        \toprule
        Method & $\tau_a$ & mAP & Rec & Prec & Acc & F1 \\
        \midrule
        STTran~\cite{cong2021spatial} 
        & - & 28.32 & - & - & - & - \\
        ST-Gaze*~\cite{ni2023human} 
        & - & 38.46 & \textbf{73.62} & 59.16 & \textbf{53.76} & 60.57 \\
        \midrule        
        \multirow{3}{*}{SAGE (Ours)} 
        & 1 & \textbf{38.65} & 72.44 & 59.22 & 52.22 & 61.96 \\
        & 3 & 38.18 & 72.21 & \textbf{59.95} & 52.18 & \textbf{62.02}  \\
        & 5 & 38.13 & 72.01 & 59.92 & 52.18 & 61.88 \\
        \bottomrule
    \end{tabular}}
    \label{table:HOI Vid detection}
\end{minipage}
\hfill
\begin{minipage}[t]{0.48\textwidth}
    \centering
    \setlength{\tabcolsep}{1pt}
    \caption{\textbf{HOI anticipation} in Oracle mode on Vid-HOI dataset. * includes word embedding module.}
    \scalebox{0.8}{
    \begin{tabular}{c c c c c c c}
        \toprule
        Method & $\tau_a$ & mAP & Rec & Prec & Acc & F1 \\
        \midrule
        \multirow{3}{*}{STTran~\cite{cong2021spatial}} 
        & 1 & 29.09 & \textbf{74.76} & 41.36 & 36.61 & 50.48 \\
        & 3 & 27.59 & \textbf{74.79} & 40.86 & 36.42 & 50.16 \\
        & 5 & 27.32 & \textbf{75.65} & 41.18 & 36.92 & 50.66 \\
        \multirow{3}{*}{ST-Gaze*~\cite{ni2023human}} 
        & 1 & 35.71 & 71.28 & 59.38 & 51.06 & 62.06 \\
        & 3 & 32.19 & 71.09 & 60.17 & 51.63 & 62.39 \\
        & 5 & 32.30 & 70.67 & 58.99 & 50.79 & 61.59 \\
        \midrule
        \multirow{3}{*}{SAGE (Ours)} 
        & 1 & \textbf{37.71} & 72.88 & \textbf{60.24} & \textbf{51.86} & \textbf{62.72} \\
        & 3 & \textbf{34.22} & 72.36 & \textbf{60.98} & \textbf{52.88} & \textbf{63.05} \\
        & 5 & \textbf{32.64} & 71.96 & \textbf{59.88} & \textbf{51.46} & \textbf{62.14} \\
        \bottomrule
    \end{tabular}}
    \label{table:HOI Vid anticipation}
\end{minipage}
\vspace{-1em}
\end{table*}

\begin{table}[t]
\small
\centering
\setlength{\tabcolsep}{1.5pt}
\caption{\textbf{SAGE results on EGTEA Gaze+.} $^\dagger$adapted for gaze anticipation.}
\scalebox{0.80}{
{
\begin{tabular}{c|ccc|ccc|ccc|c}
\toprule
\multirow{2}{*}{Method}                                                                          & \multicolumn{3}{c|}{\begin{tabular}[c]{@{}c@{}}Gaze \\ Detection\end{tabular}}                                                          & \multicolumn{3}{c|}{\begin{tabular}[c]{@{}c@{}}Action \\ Recognition\end{tabular}}                                                                                                       & \multicolumn{3}{c|}{\begin{tabular}[c]{@{}c@{}}Gaze \\ Anticipation\end{tabular}}                                                       & \begin{tabular}[c]{@{}c@{}}Action \\ Anticipation\end{tabular} \\ \cline{2-11} & F1 & Rec& Prec& S1 & S3 (Top-1) & Avg. (Top-1) & F1  & Rec  & Prec  &  Mean-cls Acc. \\ \hline
I3D-R50~\cite{feichtenhofer2019slowfast}  & 40.9  & 57.2   & 31.8           & -  & -   & -     & -  & - & - & -\\
MViT~\cite{fan2021multiscale} & 43.0   & 57.8   & 35.4 & -      & -  & -   & - & - & - & - \\
GLC~\cite{GLC} & 44.8 & 61.2 & 35.3 & -  & -    & -  & -    & - & - & - \\
I3D-2Stream~\cite{li2021eye} & - & - & - & - & 53.6 & 54.2 & - & - & -  \\
R34-2Stream~\cite{sudhakaran2018attention} & -  & - & -& - & 58.6 & 60.8 & - & - & - & - \\
SAP~\cite{wang2020symbiotic} & - & - & - & - & 62.0 & 62.7  & - & - & - & - \\
GC-TSM~\cite{hao2022group} & - & - & - & - & 62.6  & \textbf{65.1} & - & - & - & - \\
I3D-R50~\cite{feichtenhofer2019slowfast}$^\dagger$ & - & - & - & -& -  & - &  34.3  & 46.5  & 29.6  & - \\
MViT~\cite{fan2021multiscale}$^\dagger$  & - & - & - & - & - & - &   31.5 & 44.8 & 28.5  & - \\
CSTS-Visual~\cite{lai2024listen}  & - & - & - & -& - & -  &        31.3 & 46.2 & 28.1  & - \\                                        
GLC~\cite{GLC}$^\dagger$ & - & - & - & - & - & - & 32.8 & 48.5 & 28.7 & - \\
AFFT~\cite{AFFT}& - & - & - & -& - & - & - & - & - & 35.2 \\
AVT~\cite{AVT}  & - & - & - & -& - & -  &  - & - & - & 35.2 \\        MF~\cite{patrick2021keeping} & - & - & - & -& -  & - & - & - & - & 56.9  \\
ORVIT-MF~\cite{herzig2022object} & - & - & - & - & - & - & - & - & - & 57.2 \\                                                         
InAViT~\cite{InAViT} & - & - & - & -& - & -  & - & - & - & 58.2                                                          \\ \hline

MCN~\cite{huang2020mutual} & - & - & - & 55.7 & - & - & - & - & - & - \\
EGTEA~\cite{li2021eye} & 33.84 & 41.93 & 28.41 & 57.2 & - & -  & - & - & - & - \\ \hline
{\color[HTML]{C0C0C0} TSF-B~\cite{zhao2023learning}}  & {\color[HTML]{C0C0C0} -} & {\color[HTML]{C0C0C0} -} & {\color[HTML]{C0C0C0} -} & {\color[HTML]{C0C0C0} -} & {\color[HTML]{C0C0C0} -}& {\color[HTML]{C0C0C0} 65.6} & {\color[HTML]{C0C0C0} -}  & {\color[HTML]{C0C0C0} -}  & {\color[HTML]{C0C0C0} -}  & {\color[HTML]{C0C0C0} - } \\

{\color[HTML]{C0C0C0} Ego-VPA~\cite{wu2025ego}}  & {\color[HTML]{C0C0C0} -} & {\color[HTML]{C0C0C0} -} &  {\color[HTML]{C0C0C0} -} & {\color[HTML]{C0C0C0} -} & {\color[HTML]{C0C0C0} -} & {\color[HTML]{C0C0C0} 73.4} & {\color[HTML]{C0C0C0} -}  & {\color[HTML]{C0C0C0} -}  & {\color[HTML]{C0C0C0} -}  & {\color[HTML]{C0C0C0} - } \\ 
{\color[HTML]{C0C0C0} LaViLa-L~\cite{pei2025modeling}}  & {\color[HTML]{C0C0C0} -} & {\color[HTML]{C0C0C0} -} & {\color[HTML]{C0C0C0} -} & {\color[HTML]{C0C0C0} -} & {\color[HTML]{C0C0C0} -} & {\color[HTML]{C0C0C0} 76.0} & {\color[HTML]{C0C0C0} -}  & {\color[HTML]{C0C0C0} -}  & {\color[HTML]{C0C0C0} -}  & {\color[HTML]{C0C0C0} - } \\

{\color[HTML]{C0C0C0} LaViLa~\cite{zhao2023learning}}  & {\color[HTML]{C0C0C0} -} & {\color[HTML]{C0C0C0} -} & {\color[HTML]{C0C0C0} -} & {\color[HTML]{C0C0C0} -} & {\color[HTML]{C0C0C0} -} &  {\color[HTML]{C0C0C0} 81.8} & {\color[HTML]{C0C0C0} -}  & {\color[HTML]{C0C0C0} -}  & {\color[HTML]{C0C0C0} -}  & {\color[HTML]{C0C0C0} - } \\ 

{\color[HTML]{C0C0C0} EgoVideo-G~\cite{pei2025modeling}} & {\color[HTML]{C0C0C0} -} & {\color[HTML]{C0C0C0} -} & {\color[HTML]{C0C0C0} -} & {\color[HTML]{C0C0C0} -}  & {\color[HTML]{C0C0C0} -} & {\color[HTML]{C0C0C0} 80.0} & {\color[HTML]{C0C0C0} -}  & {\color[HTML]{C0C0C0} -}  & {\color[HTML]{C0C0C0} -}  & {\color[HTML]{C0C0C0} - } \\ \hline
SAGE (Ours)  & \textbf{46.8}   & \textbf{62.1} & \textbf{36.8} & \textbf{61.8} & \textbf{63.1}  & 65.0    & \textbf{37.0}  & \textbf{54.9}   & \textbf{32.7}  & \textbf{58.4}\\
\bottomrule
\end{tabular}}
}
\vspace{-1em}
\label{table: SAGE on EGTEA}
\end{table}


\subsection{SAGE on Egocentric Videos}\label{section:exp-ego}
We conduct a comprehensive evaluation on the EGTEA Gaze+ dataset~\cite{EGTEAGaze+} in Table~\ref{table: SAGE on EGTEA}, covering four tasks as described below. For the anticipation setup, we use the standard horizon of $\tau_a=0.5s$ and report results for additional horizons in section~\ref{supp anticipation horizon}. \\
\noindent\textbf{Gaze Detection \& Anticipation.}
We compare our gaze estimation performance against existing egocentric gaze estimation models, including GLC~\cite{GLC}, MViT~\cite{fan2021multiscale}, and I3D-R50~\cite{feichtenhofer2019slowfast}. As shown in Table~\ref{table: SAGE on EGTEA}, SAGE achieves the best F1 score of 46.8, significantly surpassing GLC (44.8), MViT (43.0), and I3D-R50 (40.9). It also achieves the highest recall at 62.1 and the top precision at 36.8. We improve GLC with 0.9 and 1.5 gain in recall and precision, respectively. 
For gaze anticipation task, as no existing baseline exists on EGTEA Gaze+, we established four baseline models by re-training GLC, MViT, I3D-R50 and CSTS-Visual~\cite{lai2024listen}) with ground-truth future gaze heatmaps.
In the sixth row of Table~\ref{table: SAGE on EGTEA}, our method achieves the higher F1 score (37.0), recall (54.9), and precision (32.7) than four of the baseline models.
These results highlight the advantage of integrating gaze modeling with action understanding under joint training framework. SAGE can serve as a strong baseline model for gaze anticipation on EGTEA Gaze+. We also study the sensitivity of SAGE to gaze modules in section~\ref{supp: section egtea}.\\
\noindent\textbf{Action Recognition.}
In Table~\ref{table: SAGE on EGTEA}, we compare SAGE performances with other state-of-the-art methods on the EGTEA Gaze+ dataset in terms of average accuracy across three test splits. On the third test set (S3), SAGE achieves an accuracy of 63.1, surpassing the performance of GC-TSM~\cite{hao2022group}, which reports 62.6. Overall, SAGE obtains an average accuracy of 65.0, which is comparable to GC-TSM’s top score of 65.1, while outperforming other recent baselines such as SAP (62.7) and R34-2Stream (60.8). Since SAGE does not leverage foundation models (LLMs or VLMs) or large-scale external pretraining, we exclude recent models (e.g., LaViLa~\cite{zhao2023learning}, Ego-VPA~\cite{wu2025ego}, EgoVideo~\cite{pei2025modeling} etc) from direct comparison, but still report it in Table~\ref{table: SAGE on EGTEA} for completeness. Furthermore, unlike SAGE, these approaches are limited to action recognition and do not report results on additional tasks. \\
\noindent\textbf{Action Anticipation.}
The full SAGE model outperforms prior models including AVT~\cite{AVT}, and InAViT~\cite{InAViT}, achieving the highest mean-class accuracy of 58.4, surpassing the latest work InAViT (58.2). Compared to transformer-based methods such as MF (56.9) and AVT (35.2), SAGE demonstrates a clear advantage in capturing gaze-conditioned temporal dependencies for anticipation.
We show qualitative results on EGTEA Gaze+ in Figure~\ref{fig:vis_combined}.

\subsection{Action Uncertainty from Gaze}
\label{main section: uncertainty}
\begin{figure}[t!]
    \centering
    \includegraphics[width=0.92\textwidth, height=0.24\textwidth]{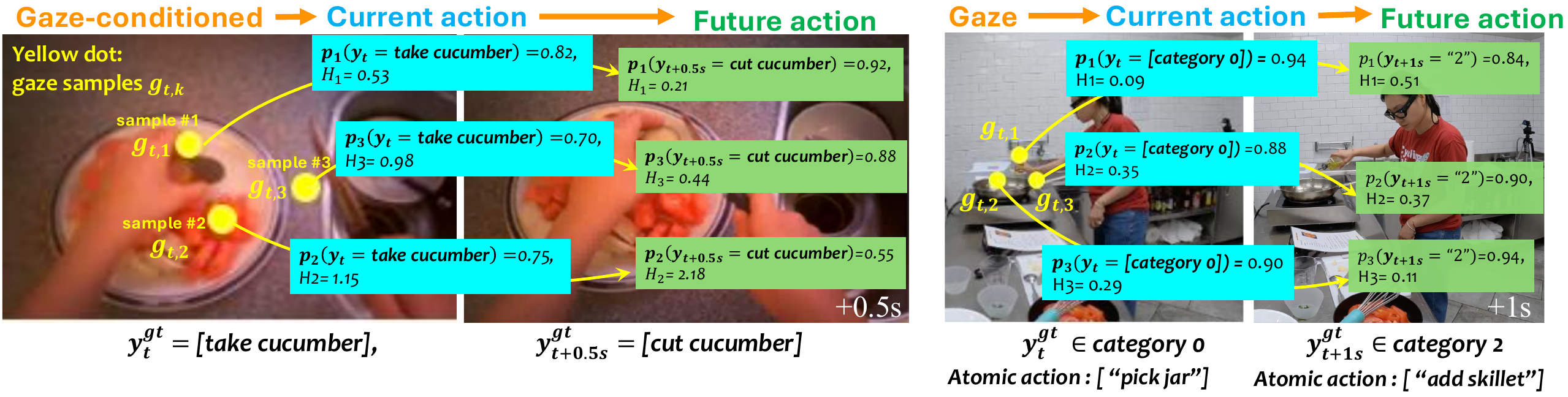}
    \caption{\textbf{Gaze-conditioned action uncertainty.} We show two examples of uncertainty quantification for current action $y_t$ and future action $y_{t+M}$. Sampling three gaze locations produces varying probabilities and entropies: in EGTEA Gaze+ (left), both current and future HOIs are gaze-sensitive, whereas in Exo-Cook predictions remain relatively stable, indicating robust action inference.}
    \label{fig:action uncertainty}
\vspace{-4mm}
\end{figure}

The first work to model uncertainty in gaze measurements for action recognition is \cite{li2021eye}. However, \cite{li2021eye} considers only uncertainty in the gaze distribution, while uncertainty in action prediction is not explicitly modeled. Moreover, the method focuses solely on uncertainty of gaze for estimating the current action and does not model how it influences future action predictions. 
In contrast, SAGE quantifies uncertainty for both current and future actions. As shown in Figure~\ref{fig:overview} (b), gaze locations are sampled through Monte Carlo sampling and propagate to action detection and anticipation through GCSA and GCTP module. Action uncertainty can be computed by marginalizing over the gaze samples. Given $K$ Monte Carlo gaze samples $\{g^{k}\}_{k=1}^K$ from predicted gaze heatmap, we obtain action distributions $\mathbf{p}^{k} = p(y \mid g^{k}, I)$ and compute conditional entropy $H(y \mid I)$ and mutual information $\mathrm{MI}(y,g\mid I)$ for action.
(1) The predictive distribution is approximated as
$\bar{\mathbf{p}} = \frac{1}{K}\sum_{k=1}^{K} \mathbf{p}^{k}$. The total predictive uncertainty is measured by the entropy: $H(y \mid I) = -\sum_c \bar{p}_c \log \bar{p}_c$.
(2) The mutual information (MI) of HOI predictions over gaze, represents how sensitive HOI prediction is to gaze uncertainty: $\mathbb{E}[H] = \frac{1}{K}\sum_{k=1}^{K}
(-\sum_c p_c^{k} \log p_c^{k})$, $\mathrm{MI}(y,g\mid I) = H(y \mid I) - \mathbb{E}[H].$
In Figure~\ref{fig:action uncertainty}, we provide two examples of sampling multiple gaze location from gaze heatmap and obtain corresponding gaze-conditioned uncertainty $H_y$ for current action $y_t$ and future action $y_{t+M}$. In the EGTEA Gaze+ example, small changes in gaze location lead to  variations in both prediction confidence and entropy for the current and future actions. In contrast, the Exo-Cook example exhibits substantially lower action sensitivity to gaze. We show more uncertainty quantification details in section~\ref{supp action uncertainty}.

\section{Conclusion}
We presented SAGE, a unified end-to-end framework for simultaneous recognition and anticipation of human-object interactions and gaze. By introducing the proposed GCSA and GCTP modules, SAGE explicitly models the synergy between gaze and actions across both current and future time horizons, enabling more comprehensive human behavior understanding. Its modular design supports both egocentric and exocentric settings, and is validated on VidHOI, EGTEA Gaze+, and our new Exo-Cook benchmark. Extensive experiments demonstrate that jointly modeling gaze and actions consistently matches or surpasses specialized state-of-the-art methods across multiple tasks. By unifying human attention and actions within a single framework, SAGE provides a promising foundation for future research in human behavior analysis and intuitive human–machine interaction.
\clearpage
\appendix
\section{Supplementary}
\subsection{Implementation Details} \label{supp implementation details}
\subsubsection{Human-object-Pair Feature Extraction} \label{supp implementation details hoi extraction}
We refer to ST-Gaze~\cite{ni2023human} to pre-process Vid-HOI and Exo-Cooking datasets. 
An object module is applied to process the video sequence
and detects $N$ bounding boxes \(\{b_{t,j}\}\) along with their corresponding classes. Human bounding boxes are also detected. In addition, human head are detected by YOLO-v5 model \cite{jocher2022yolov5} and paired with their body bounding boxes. DeeoSort model is applied to associate these detections with previous detections to establish trajectories for each detected human \(\{H_i\}\) and object \(\{O_j\}\).
Video feature are extracted by ResNet.
\subsubsection{Viewpoint adaptability} As mentioned in main paper (section 4.1), SAGE can adapt to egocentric and exocentric viewpoint with differing feature construction process. Egocentric (FPV) and exocentric (TPV) scenarios have typically been studied separately in gaze and action understanding, as their visual characteristics differ substantially and are rarely addressed within a single model. Consequently, prior work does not report results across both settings. To the best of our knowledge, SAGE is the first framework that supports both FPV and TPV within a unified architecture for four tasks: gaze detection, HOI detection, gaze anticipation, and HOI anticipation. The framework adapts viewpoint-specific modalities during feature encoding and gaze estimation while maintaining a consistent network flow and output representation. Note that this work does not claim cross-view generalization between FPV and TPV domains.

During feature encoding, the key difference between FPV and TPV arises in the construction of human–object interaction tokens. For each frame $t$, we extract human and object bounding boxes and construct human–object pair tokens
\[
\mathbf{x}^o_t =
[\mathbf{x}_{t,\langle 1,1 \rangle}, \dots, \mathbf{x}_{t,\langle i,j \rangle}, \dots, \mathbf{x}_{t,\langle n_t^h, n_t^o \rangle}],
\]
where $n_t^h$ and $n_t^o$ denote the numbers of detected humans and objects. Each token $\mathbf{x}_{t,\langle i,j \rangle}$ encodes the interaction between human $i$ and object $j$, including appearance features and spatial relationships. 
The representation of $\mathbf{x}_{t,\langle i,j \rangle}$ differs slightly between viewpoints. In TPV videos, multiple subjects may appear in the scene ($i \geq 1$), and the token encodes scene appearance, human body features, body location, and the spatial relationship between humans and objects. These cues are important for recognizing interactions from a third-person perspective. In contrast, FPV videos lack explicit global body location because the camera is attached to the actor. Instead, interaction tokens encode the spatial relationships between detected hands and objects, with the human index fixed to $n_t^h=1$.
Another difference arises in gaze estimation. In TPV videos, SAGE predicts gaze using head appearance and orientation features following prior work such as~\cite{chong2020detecting}. In FPV videos, gaze is interpreted as an attention region within the egocentric scene, and the gaze model predicts a spatial attention distribution from scene features. Despite these viewpoint-specific adaptations, the subsequent gaze-conditioned reasoning modules (GCSA and GCTP) remain identical across FPV and TPV, enabling a unified framework for joint gaze–action modeling.

\subsubsection{Model Architecture}
SAGE jointly performs four tasks and the detailed SAGE architecture can be divided into four parts.

\noindent\textbf{SAGE-1 for Gaze Detection.}
We integrate existing gaze detection models into SAGE framework to produce differentiable gaze heatmaps. The gaze detection module comprises a visual encoder and a decoder that outputs gaze fixation heatmaps. The input sequence is divided into non-overlapping patches and each patch is flattened and projected into a $D$-dimensional embedding space via a linear mapping. These video tokens are then processed by transformer layers composed of multiple self-attention blocks. To generate gaze heatmaps, a transformer decoder—based on multiscale self-attention blocks from MViT~\cite{fan2021multiscale}—upsamples the encoded features. The decoder outputs feature maps of shape $T'H'W' \times D'$, followed by a softmax operation over the last dimension to obtain the final gaze heatmap $\bm{g}_{0:t}$. For the egocentric setting, we adopt the GLC~\cite{GLC} model as our gaze detection architecture. For exocentric videos, we use the architecture proposed by~\cite{chong2020detecting}, initializing our model with their pretrained weights.

\noindent\textbf{SAGE-2 for HOI Detection.}
Let $\mathbf{X}_t \in \mathbb{R}^{(N+1)\times D}$ denote the input tokens at frame $t$, consisting of $N$ human--object pair tokens and one global token, where $D$ is the model dimension. The spatial encoder contains $N_d$ Transformer layers. Let $\mathbf{H}_t^{(0)}=\mathbf{X}_t$. For layer $l=1,\dots,N_d$, the projections are
\begin{equation}
Q_t^{l} = \mathbf{H}_t^{(l-1)} W_Q^{l}, \quad
K_t^{l} = \mathbf{H}_t^{(l-1)} W_K^{l}, \quad
V_t^{l} = \mathbf{H}_t^{(l-1)} W_V^{l},
\end{equation}
where $W_Q^{l}, W_K^{l}, W_V^{l} \in \mathbb{R}^{D\times D}$. For $H$ attention heads, each head has dimension $d_k = D/H$. The gaze bias matrix is $\mathbf{S}_t$. Each Transformer layer updates the token representations with the GCSA operation:
\begin{equation}
\mathrm{GCSA}(Q_t^{l}, K_t^{l}, V_t^{l}, \mathbf{S}_t)
=
\mathrm{Softmax}\!\left(
\frac{Q_t^{l} (K_t^{l})^\top + \mathbf{S}_t}{\sqrt{d_k}}
\right)V_t^{l} .
\end{equation}
\begin{align}
\widetilde{\mathbf{H}}_t^{l}
&=
\mathrm{LN}\!\left(
\mathbf{H}_t^{(l-1)} +
\mathrm{GCSA}(Q_t^{l}, K_t^{l}, V_t^{l}, \mathbf{S}_t)
\right), \\
\mathbf{H}_t^{l}
&=
\mathrm{LN}\!\left(
\widetilde{\mathbf{H}}_t^{l} +
\mathrm{FFN}^{l}(\widetilde{\mathbf{H}}_t^{l})
\right),
\end{align}
where $\mathrm{FFN}^{l}: \mathbb{R}^{D} \rightarrow \mathbb{R}^{D_{\mathrm{ff}}} \rightarrow \mathbb{R}^{D}$ is a feed-forward network. After $N_d$ layers, the spatial encoder outputs $\mathbf{C}_t = \mathbf{H}_t^{(N_d)} \in \mathbb{R}^{(N+1)\times D}$.
The frame-level gaze-aware HOI feature is obtained from the global token $\mathbf{c}_t \in \mathbb{R}^{D}$.
Given the sequence of frame features
$[\mathbf{c}_0, \dots, \mathbf{c}_t] \in \mathbb{R}^{(t+1)\times D}$,
we apply a temporal Transformer encoder with one layer

\begin{equation}
\mathbf{z}_{0:t}
=
\mathrm{TransformerEnc}_{1}([\mathbf{c}_0, \dots, \mathbf{c}_t])
\in \mathbb{R}^{(t+1)\times D}.
\end{equation}
Let $\mathbf{z}_t \in \mathbb{R}^{D}$ denote the final token corresponding to time $t$. The HOI detection head is an MLP $\phi: \mathbb{R}^{D} \rightarrow \mathbb{R}^{C_y}$, where $C_y$ is the number of HOI classes. The action probability is predicted as
\begin{equation}
p(y_t \mid g_{0:t}, I_{0:t})
=
\mathrm{Softmax}\!\left(\phi(\mathbf{z}_t)\right).
\end{equation}

\noindent\textbf{SAGE-3 for Gaze Anticipation.}
Each gaze heatmap $g_k \in \mathbb{R}^{64\times64}$ is embedded into a latent token with periodic position encoding (PPE) \cite{fan2022faceformer} : $\tilde{\mathbf{g}}_k =
\mathrm{Conv}(g_k) + \mathrm{PPE}(k),$
where $\tilde{\mathbf{g}}_k \in \mathbb{R}^{D}$, $D$ is the model dimension and $PPE(\cdot)$ is denoted as:
\begin{equation}
\begin{aligned}
    & PPE_{(t,2i)} = \sin((t \text{ mod } P) / (10000)^{2t/d}) \\
    & PPE_{(t,2i+1)} = \cos((t \text{ mod } P) / (10000)^{2t/d})
\end{aligned}
\label{eq:PPE}
\end{equation}
Stacking all tokens yields the gaze sequence $
[\tilde{\mathbf{g}}_0,\dots,\tilde{\mathbf{g}}_t]
\in \mathbb{R}^{(t+1)\times D}.$
We use a Transformer encoder with $N_g$ layers to model the temporal dependencies of gaze trajectories:
\begin{equation}
\mathbf{h}^g_{0:t}
=
\mathrm{TransformerEnc}_{N_g}([\tilde{\mathbf{g}}_0,\dots,\tilde{\mathbf{g}}_t])
\in \mathbb{R}^{(t+1)\times D}.
\end{equation}
Let $\mathbf{X}^o_{0:t}\in\mathbb{R}^{(t+1)\times N \times D}$ denote the human–object interaction tokens, $\mathbf{H}^g = \{ \mathbf{h}^g_{0:t} \}$.
A Cross-attention layer is applied to integrate video interaction context with gaze features: 
\begin{equation}
\mathbf{z}^g_{0:t}
=
\mathrm{MHCA}(\mathbf{H}^g,\mathbf{X}^o_{0:t})
\in\mathbb{R}^{(t+1)\times D}.
\end{equation}
The future gaze heatmaps are generated by a decoder
\begin{equation}
\hat{g}_{m}
=
\mathcal{D}_{\phi}(\mathbf{z}^g_t),
\quad m=t+1,\dots,t+M,
\end{equation}

\noindent\textbf{SAGE-4 for HOI Anticipation.} 
Let $\hat{\mathbf{G}} = \{\bm{\hat{\mathbf{g}}}_{t+1:t+M}\} \in \mathbb{R}^{M \times D}$ denote the embedded future gaze tokens, where $M$ is the prediction horizon and $D$ is the model dimension. Temporal correlations among future gaze shifts are modeled by a MHSA layer.
For the $l$-th layer, the projections are
\begin{equation}
Q^{l} = \hat{\mathbf{G}}^{(l-1)} W_Q^{l}, \quad
K^{l} = \hat{\mathbf{G}}^{(l-1)} W_K^{l}, \quad
V^{l} = \hat{\mathbf{G}}^{(l-1)} W_V^{l},
\end{equation}
where $W_Q^{l}, W_K^{l}, W_V^{l} \in \mathbb{R}^{D \times D}$. For $H$ attention heads, each head has dimension $d_k = D/H$. The self-attention output is

\begin{equation}
\mathrm{MHSA}(\hat{\mathbf{G}}^{(l-1)}) =
\mathrm{Softmax}\!\left(
\frac{Q^{l} (K^{l})^\top}{\sqrt{d_k}}
\right)V^{l}.
\end{equation}
The historical temporal gaze representation is $\mathbf{A}^{l} =
\mathrm{LN}\!\left(
\hat{\mathbf{G}}^{(l-1)} +
\mathrm{MHSA}(\hat{\mathbf{G}}^{(l-1)})
\right).$
To condition future HOI prediction on the current interaction context, we apply cross-attention between the gaze tokens and the current gaze-conditioned HOI feature $\mathbf{c}_t \in \mathbb{R}^{D}$, with $Q_c = \mathbf{A}^{l} W_Q^c, \quad
K_c = \mathbf{c}_t W_K^c, \quad
V_c = \mathbf{c}_t W_V^c,$ The MHCA is calculated as:
\begin{equation}
\mathrm{MHCA}(\mathbf{A}^{l}, \mathbf{c}_t)
=
\mathrm{Softmax}\!\left(
\frac{Q_c K_c^\top}{\sqrt{d_k}}
\right)V_c ; \ 
\hat{\mathbf{A}}^{l} =
\mathrm{LN}\!\left(
\mathbf{A}^{l} +
\mathrm{MHCA}(\mathbf{A}^{l}, \mathbf{c}_t)
\right).
\end{equation}
In conclusion, the gaze-conditioned temporal representation (matrix $\bm{\hat{\mathbf{a}}}_{t+1:t+M}$ denoted as $\hat{\mathbf{A}}$) for future HOI feature is generated from
\begin{equation}
\hat{\mathbf{A}}
=
\mathrm{TransformerEnc}_{N_a}(\hat{\mathbf{G}}, \mathbf{c}_t)
\in \mathbb{R}^{M \times D}.
\end{equation}
and the anticipated HOI at step $t+M$ is predicted by
\begin{equation}
\hat{y}_{t+M}
=
\mathrm{Softmax}(\mathrm{MLP}(\hat{\mathbf{a}}_{t+M})).
\end{equation}
\subsubsection{Monte Carlo Gaze Sampling} \label{supp: monte carlo sampling}
To propagate gaze uncertainty into both current and future action prediction, we approximate the marginal likelihood in Eq.1 (main paper) using Monte Carlo sampling over gaze trajectories.
The gaze detector predicts a heatmap for each observed frame independently. Let $\hat{g}_\tau \in \mathbb{R}^{H_g \times W_g}$ denote the predicted gaze heatmap at time $\tau$, where $\tau=0,\dots,t$. We first normalize each heatmap into a spatial distribution
\begin{equation}
p_\tau(h,w) = \frac{\hat{g}_\tau(h,w)}{\sum_{h',w'} \hat{g}_\tau(h',w')}.
\end{equation}
Assuming conditional independence across observed frames, the present gaze distribution is
\begin{equation}
p(\bm{g}_{0:t}\mid I_{0:t}) = \prod_{\tau=0}^{t} p(g_\tau \mid I_\tau).
\end{equation}
For each Monte Carlo sample $k\in\{1,\dots,K\}$, we independently sample one gaze fixation location from each frame-wise distribution,
The sampled present gaze trajectory is then
\begin{equation}
\bm{g}_{0:t}^{k} = \big(g_0^{k}, g_1^{k}, \dots, g_t^{k}\big).
\label{supp:eq MC gaze current}
\end{equation}
Conditioned on the sampled present gaze trajectory $\bm{g}_{0:t}^{k}$ and observed video features, the gaze anticipation module predicts future gaze heatmaps
\begin{equation}
\hat{\bm{g}}_{t+1:t+M}^{k} \sim p(\bm{g}_{t+1:t+M}\mid \bm{g}_{0:t}^{k}, I_{0:t}).
\label{supp:eq MC gaze future}
\end{equation}
For each future step $t+m$, we again normalize the predicted heatmap into a 2D gaussian distribution, sample one gaze location.
Combining Eq~\ref{supp:eq MC gaze current} and Eq~\ref{supp:eq MC gaze future} yields a full sampled gaze trajectory
\begin{equation}
G^{k} = \big(\bm{g}_{0:t}^{k}, \bm{g}_{t+1:t+M}^{k}\big).
\end{equation}
For each sampled trajectory $G^{k}$, the current and future action probabilities are evaluated as
\begin{equation}
f(G^{k}) =
p(y_t \mid \bm{g}_{0:t}^{k}, I_{0:t}) \;
p(\bm{y}_{t+1:t+M} \mid y_t, \bm{g}_{t+1:t+M}^{k}, I_{0:t}).
\end{equation}
The marginal likelihood is approximated by
\begin{equation}
\widehat{p}(y_{t:t+M}\mid I_{0:t})
=
\frac{1}{K}\sum_{k=1}^{K} f(G^{k}).
\label{eq:mc_training}
\end{equation}
During training, since the gaze detection model is loaded from a pre-trained model, we use a small number of samples (\(K=1\)) for efficiency. During inference, we can sample multiple gaze trajectories ($K \geq 2$ )for analyzing action uncertainty conditioning on gaze.
\subsubsection{Training Details}
Based on the complexity of the full SAGE model and the difficulty in training for all the tasks. We initialize some models with pre-trained model and multi-stage training process.
\begin{itemize}
    \item The Gaze detection module is initialized with the pre-trained model from VideoAttn~\cite{chong2020detecting} (on Vid-HOI dataset) or GLC ~\cite{GLC}. 
    \item We first train joint model SAGE-12 for 5 epochs with gaze labels and action labels for the current sequence. 
    \item Then we load the weights from SAGE-12 and train the full model SAGE with future gaze and action annotations. We train 25 epochs in total.
    \item In the loss function $\mathcal{L} = \lambda_1 \mathcal{L}_{hm,1} + \lambda_2 \mathcal{L}_{hm,2} + \lambda_3 \mathcal{L}_{io} + \lambda_4 \mathcal{L}_{act,1} + \lambda_5 \mathcal{L}_{act,2}$, we set $\lambda_1=1.0, \lambda_2=1.0,\lambda_3=0.5,\lambda_4=1.5,\lambda_5=1.5$. Note that $\lambda_3=0$ in egocentric case. These values were determined empirically through initial tuning experiments and remained fixed for all reported results.
\end{itemize}


\subsection{Exo-Cook Benchmark Creation} \label{supp: section exo-cook}
\begin{figure*}[ht!]
    \centering
    \includegraphics[width=\textwidth, height=2.0in]{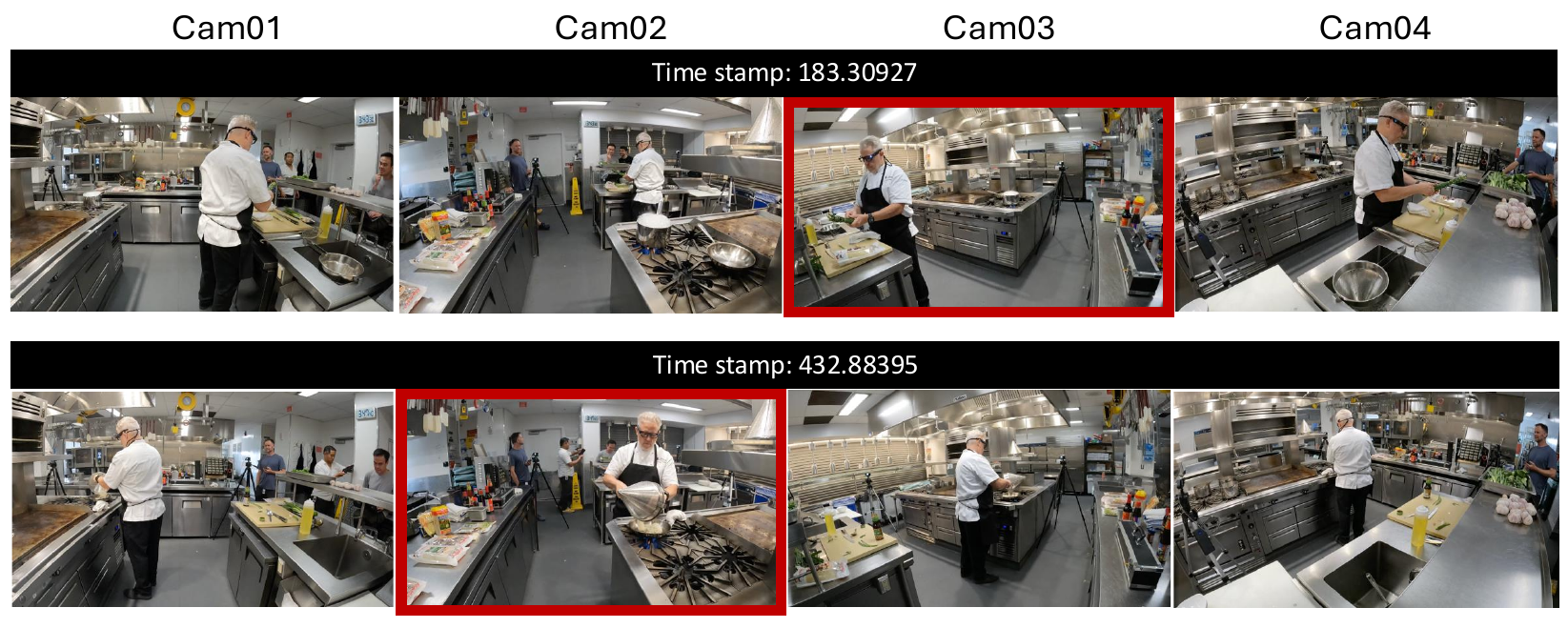}
    \caption{Visualization of the best exocentric views (based on Ego-Exo4D annotations) at different timestamps in one cooking scenario with four camera views. We show that at two different stamps (183.30927s and 432.88395s),  the best exocentric views (highlighted in red) come from Cam02 and Cam03 respectively. We generate Exo-Cook clips based on the best exocentric views.}
    \label{fig:exo-cook multiview}
\end{figure*}

We construct a benchmark based on a subset from Ego-Exo 4D. We describe more details for pre-processing and generating the labels in this section.
On cooking videos, we try BERT model \cite{devlin2019bert} to generate sentence embedding and cluster text into action categories. 
We extract the text descriptions from Ego-Exo 4D cooking videos and generate 30250 video clips and the corresponding action descriptions. 
We generate text embeddings for each video clip and apply K-means algorithm to perform action clustering based on the semantic meaning of the text. 
\subsubsection{View Diversity of Exo-Cook} 
The original Ego-Exo4D~\cite{grauman2024ego} provide multiple (4-5) exocentric videos from tripods
placed around the subject. All views are time-synchronized. Exo-Cook is sampled from Ego-Exo4D cooking videos based on the atomic action descriptions that specify what occurs at a single time point. The timestamp in the description is linked to a \texttt{best\_exo} field, which denotes the exocentric camera view that provides the clearest visual evidence of the event. Exo-Cook clips are sampled from the best exocentric capturing using the atomic action timestamp. In all, Exo-Cook includes 8,778, 8,721, 6,429, 6,253, and 1,869 clips from Cam01–Cam05 respectively. We show an example of Exo-Cook clips from different view points in Figure~\ref{fig:exo-cook multiview}. For each cooking scene, we sample 20 frames at different timestamps from the best exocentric view with valid action annotations, which will be used as the input sequence for SAGE. Each Exo-Cook clip contains a input sequence and future segments. For the anticipation task, we sample 4 frames within future 1 second. We generate annotations for each clip, including gaze heatmap, human and object bounding boxes and action labels. We introduce the annotation pipeline in the following sections.
\subsubsection{Gaze Label Creation}
\begin{figure*}[t]
    \centering
    \includegraphics[width=\textwidth, height=2.0in]{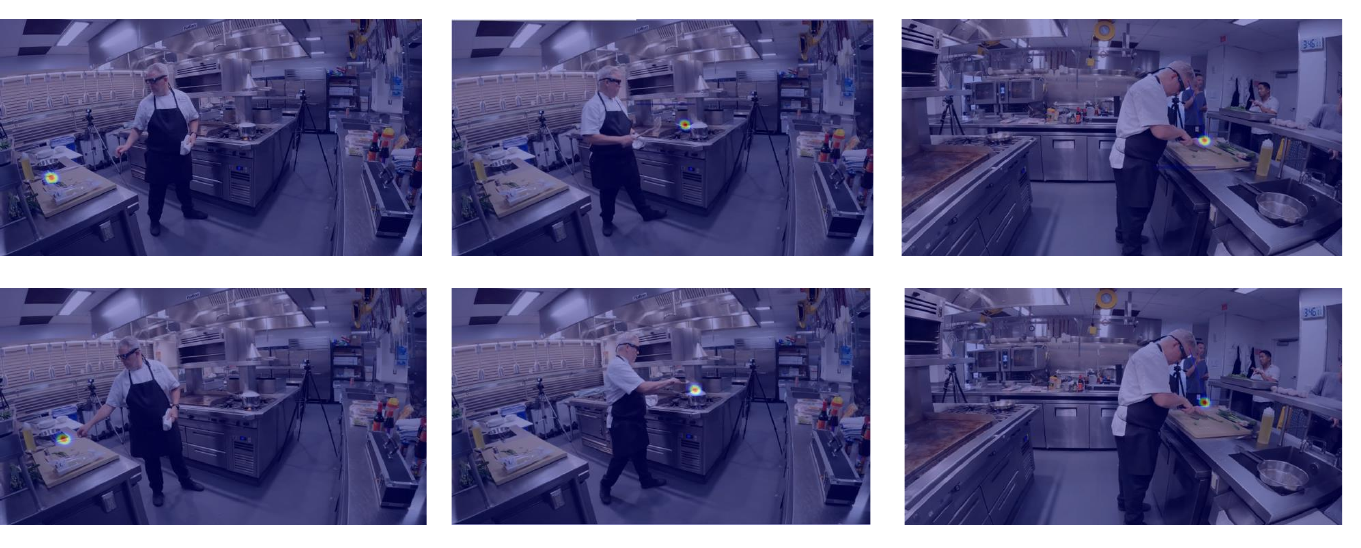}
    \caption{Individual examples (six) of created ground-truth gaze heatmap from Exo-Cook.}
    \label{fig:supp_vis_heatmap}
\end{figure*}
Ego-Exo 4D provide raw gaze annotations from the eye-tracking device (Project Aria). To generate 2D gaze labels for third-person views in Exo-Cook, we follow a calibrated projection pipeline that transforms the 3D gaze direction from the eye-tracking device (Project Aria) into 2D image coordinates of external cameras.

\noindent\textbf{Gaze Ray Extraction.}  
For each frame, we extract the 3D gaze origin and direction vectors from the left and right eyes using Aria's `scene gaze` and `eye gaze` streams. These are defined in the Aria scene camera coordinate system and describe binocular gaze rays:
\[
\text{Ray}_L = (\bm{o}_L, \bm{d}_L), \quad \text{Ray}_R = (\bm{o}_R, \bm{d}_R)
\]

\noindent\textbf{3D Gaze Estimation in Aria Frame.}  
Each frame of Project Aria includes left and right eye gaze vectors with respect to the device's coordinate frame. We estimate the 3D point of gaze $\bm{G}^{\text{aria}}_{3D}$ by computing the 3D intersection (or midpoint approximation) of the two gaze rays $\text{Ray}_L$ and $\text{Ray}_R$.

\noindent\textbf{Coordinate Transformation.}  
To localize the gaze in the third-person camera frame, we use the known extrinsic calibration between the Aria device and each external camera. Let $\bm{T}_{\text{aria} \rightarrow \text{cam}}$ denote the 6DoF pose of the Aria glasses with respect to a third-person camera. We transform the 3D gaze point into the third-person coordinate system via:
\[
\bm{G}^{\text{cam}}_{3D} = \bm{T}_{\text{aria} \rightarrow \text{cam}} \cdot \bm{G}^{\text{aria}}_{3D}
\]

\noindent\textbf{3D-to-2D Projection.}  
Given the intrinsic camera matrix $\bm{K}$ of the third-person camera and the transformed gaze point $\bm{G}^{\text{cam}}_{3D}$, we project the gaze to the image plane:
\[
\bm{G}^{\text{cam}}_{2D} = \Pi(\bm{K}, \bm{G}^{\text{cam}}_{3D})
\]
where $\Pi$ denotes the standard perspective projection. This gives us the 2D gaze fixation point in pixel coordinates.

\noindent\textbf{Heatmap Generation.}  
Following prior work~\cite{chong2020detecting}, we convert each gaze fixation point into a 2D Gaussian heatmap $\bm{M}^{\text{pseudo}} \in \mathbb{R}^{H \times W}$. The heatmap is centered at $\bm{G}^{\text{cam}}_{2D}$ with an isotropic Gaussian kernel:
\[
\sigma = \frac{W_{\text{hm}} + H_{\text{hm}}}{2} \cdot \frac{3}{64}
\]
where $W_{\text{hm}}$ and $H_{\text{hm}}$ are the dimensions of the heatmap. For a $64 \times 64$ heatmap, this yields $\sigma = 3$ pixels. We show examples of generated gaze heatmaps in Figure.~\ref{fig:supp_vis_heatmap}.

To handle occlusions or failed triangulation cases, we discard gaze points where the intersection error between eye rays exceeds a threshold. Additionally, all transformations and projections are timestamp-aligned to ensure synchronization between Aria and third-person views.

\begin{table*}[ht]
\centering
\setlength{\tabcolsep}{4pt}
\caption{Verb-object extraction and semantic object selection using spaCy and BERT. In the third column, we show the cosine similarity between the spaCy object and the object name mentioned in Exo-Cook raw annotation.}
\resizebox{0.95\textwidth}{!}{
\begin{tabular}{p{5cm} p{2cm}c p{3cm}c p{3cm}c}
\toprule
\textbf{Sentence} & \textbf{Verb Object} & \textbf{spaCy Object List} & \textbf{Object Name} \\
\midrule
C places the spoon in the bowl in his right hand. 
& places, spoon 
& spoon (0.652), bowl (1.000), hand (0.437) 
& Bowl \\
\hline
C slices the garlic on the chopping board with his right hand.
& slices, garlic 
& garlic (0.424), board (0.869), hand (0.341) 
& Chopping board \\
\hline
C adds the sliced garlic into the frying pan with his right hand.
& adds, garlic 
& garlic (0.461), pan (1.000), hand (0.378) 
& Pan \\
\hline
C adds oil into the frying pan with the oil bottle in his left hand.
& adds, oil 
& oil (0.552), pan (0.449), bottle (1.000), hand (0.437) 
& Bottle \\
\bottomrule
\end{tabular}}
\label{table:supp-verb_object_extraction}
\end{table*}

\subsubsection{Action Label Creation} \label{supp: section exo-cook label creation}
\noindent\textbf{Action Description Alignment.}
Each atomic description is associated with the following key metadata:
\begin{itemize}
    \item \texttt{start\_frame} and \texttt{end\_frame}: specifying the frame-level temporal boundaries of the described action.
    \item \texttt{source\_view}: indicating which third-person or egocentric camera the annotation corresponds to.
    \item \texttt{text}: raw textual description of the action.
\end{itemize}
To align each action with global time stamp, we use the provided frame indices and convert them to time given the video frame rate. We first sample the video frames near the \texttt{start\_frame} for action recognition. Then we define a time offset $\tau_a = 1s$ beyond the \texttt{end\_frame} to define the target for action anticipation.

\noindent\textbf{Action Simplification with spaCy.}
We use spaCy~\cite{spacy2024} to perform syntactic parsing and extract verb-object pairs from each sentence. For example, given the description \textit{“C slices the garlic on the chopping board with his right hand”}, spaCy identifies \texttt{slices} as the verb and \texttt{garlic} as the primary object. In more complex sentences containing multiple noun phrases (e.g., “C adds the sliced garlic into the frying pan with his right hand”), spaCy is used to list all candidate objects (e.g., \texttt{garlic}, \texttt{pan}, \texttt{hand}). To identify the most semantically relevant object, we compute the cosine similarity between the BERT~\cite{devlin2019bert} embedding of each (verb,noun) pair and the embedding of the full sentence. The object with the highest similarity is selected as the primary target. For instance, in the garlic example, \texttt{pan} receives the highest score, thus being identified as the object most aligned with the described action context. In Table.~\ref{table:supp-verb_object_extraction}, we show the Cosine Similarity between different sentences extracted from the annotation file of Ego-Exo 4D. 

\begin{figure*}[t!]
    \centering
    \includegraphics[width=\textwidth, height=3.2in]{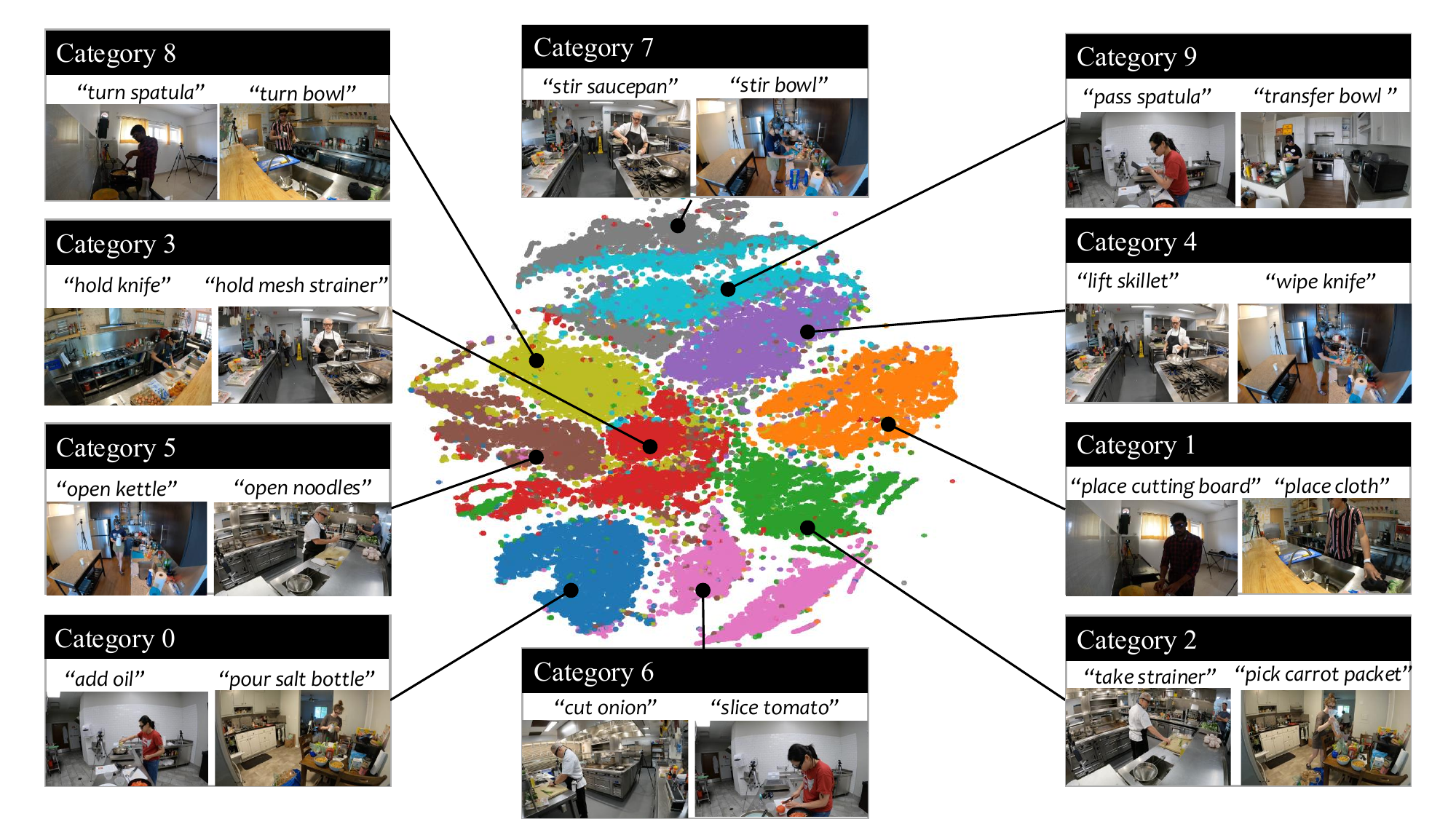}
    \caption{We illustrate the semantic interpretations of the clusters in the Exo-Cook dataset through some representative samples across 10 action categories. The clusters are obtained using K-Means on the (verb, noun) embeddings.} 
    \label{fig:vis-dataset}
\end{figure*}

\begin{figure}[t!]
    \centering
    \includegraphics[width=3.4in, height=2.2in]{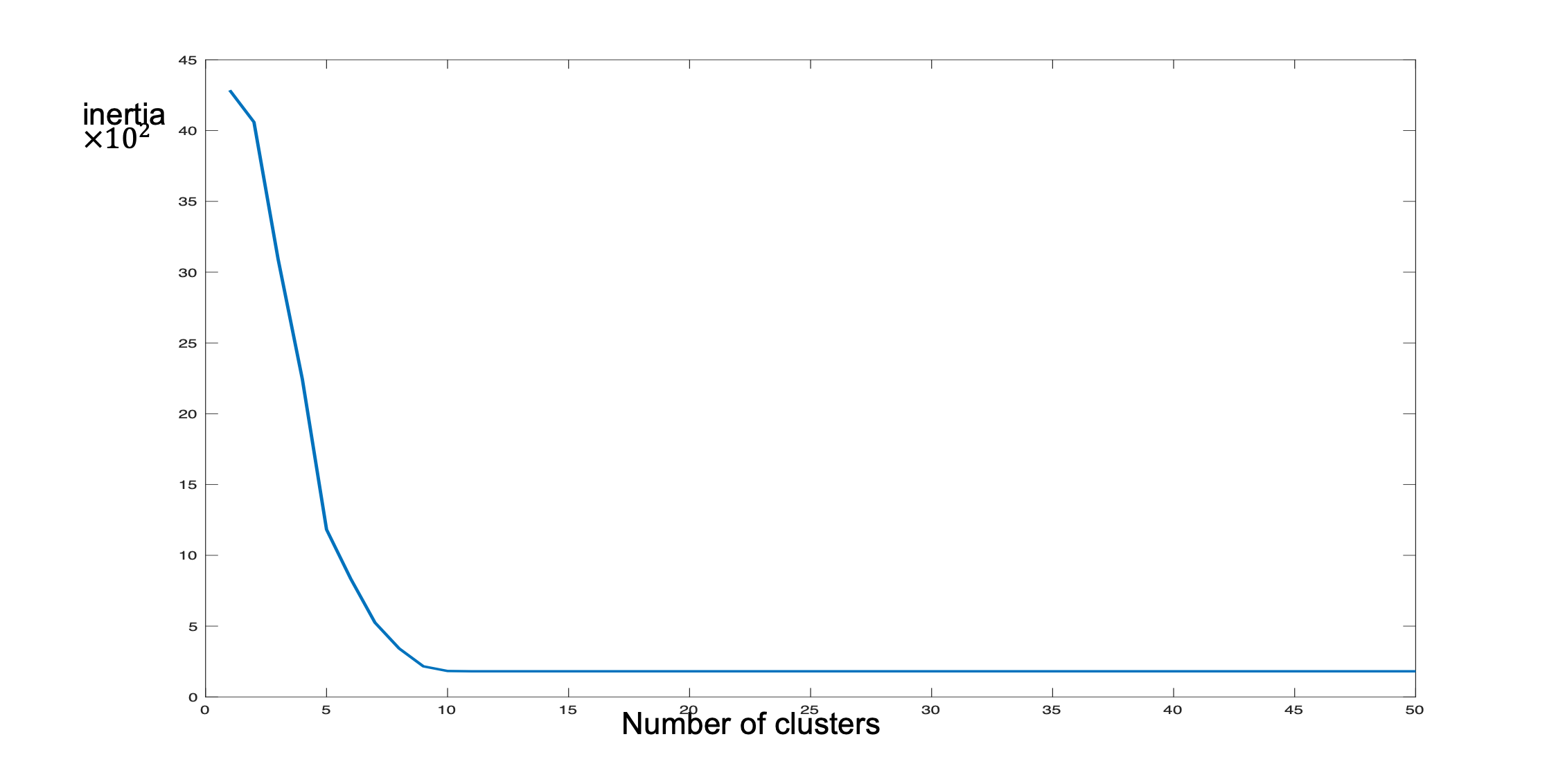}
    \caption{The inertia plot from the Elbow Method applied to all text embeddings from cooking video clips indicates the selection of $K = 10$ clusters.} 
    \label{fig:inertiaplot}
\end{figure}

\noindent\textbf{Action Clustering} As described above, we repeat the action description alignment and simplification process for all the samples we cropped from Exo-Cook.  After applying spaCy for extracting essential semantic components for each clip, we employ the pre-trained BERT~\cite{devlin2019bert} model to transform these extracted components into high-dimensional text embeddings.  
Once text embeddings for all sequences are obtained, we apply $K$-means clustering to group similar embeddings. To determine the optimal number of clusters, we use the Elbow Method by analyzing the inertia, defined as the sum of squared distances between each data point and the centroid of its assigned cluster. The inertia values across different choices of $K$ are illustrated in Figure~\ref{fig:inertiaplot}. Based on this analysis, we set the number of clusters to $K=10$ and therefore, we have 10 categories of actions in Exo-Cook. In Figure~\ref{fig:vis-dataset}, we present two representative examples from every category, accompanied by their respective (verb, noun) pairs, illustrating how each cluster category captures distinct semantic interpretations.

To further evaluate the semantic consistency of the resulting categories, we embed each action label (verb--noun pair) using a pretrained Sentence-BERT model (all-mpnet-base-v2) and compute pairwise cosine similarities. The average intra-category similarity (0.61) is substantially higher than the inter-category similarity (0.22), yielding a separation gap of 0.39 (2.77$\times$) with statistical significance ($p<10^{-4}$, permutation test).



\begin{figure*}[t!]
    \centering
    \includegraphics[width=\textwidth, height=1.8in]{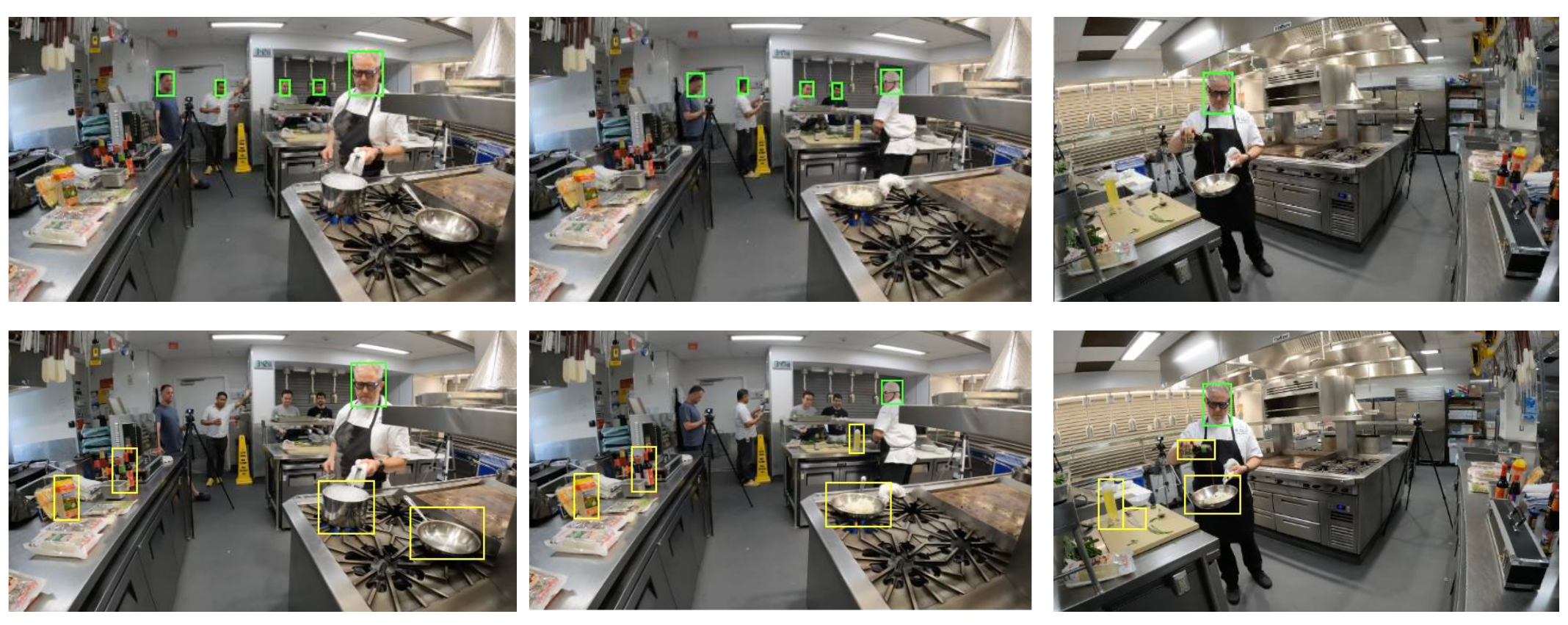}
    \caption{Examples (three) of bounding box detection for human head and object in Exo-Cook images. The first row represents head bounding box (green color) detection for all subjects in the scene. The second row shows the head bounding box for the key participant and the objct bounding boxes (yellow color).} 
    \label{fig:supp-vis-bbox}
\end{figure*}
\subsubsection{Bounding Box for Human and Object}
We follow the pipeline in~\cite{ni2023human} and use YOLOv5~\cite{jocher2022yolov5} to detect full-body and head bounding boxes of the camera wearer in third-person views. 

\noindent\textbf{Bounding Box Detection.} The first row of Figure~\ref{fig:supp-vis-bbox} illustrates examples of head bounding box detections (highlighted in green) within Exo-Cook images, capturing every person appeared in the scene. Similarly, the second row presents object bounding box detections, marked in yellow, covering all relevant objects within the scene.

\noindent\textbf{Bounding Box Matching} As our primary interest is the activity of the "major participant" equipped with the Project Aria device, we implement bounding box matching based on the projected location of the device. Specifically, the head bounding box containing the 2D projection of the Project Aria's center is designated as the major participant. Consequently, as depicted in the second row of Figure~\ref{fig:supp-vis-bbox}, any unrelated head bounding boxes will be removed. 


\subsection{Baseline Model for Gaze Anticipation}
\subsubsection{Exocentric View}
Due to lack of baseline methods on Gaze Anticipation task, we propose to adapt existing models as baseline. 
In Table~1 of the main paper, we evaluate gaze anticipation performance on the Exo-Cook dataset by adapting two baseline models: VideoAttn~\cite{chong2020detecting} and Sharingan~\cite{tafasca2024sharingan}. These baseline models, initially designed for gaze detection tasks, were modified and retrained to predict gaze heatmaps at a future timestamp (1 second ahead) based solely on the last frame of a given sequence. The results show the advantage of employing the temporal modeling in our SAGE model. 
Specifically, our temporal-aware SAGE significantly outperforms the adapted baseline models across all gaze anticipation metrics. For instance, the SAGE model (with Sharingan as its gaze detection backbone) achieves an F1 score of 49.2, notably higher than adapted VideoAttn (42.2) and adapted Sharingan (40.1). Similarly, recall (62.8 vs. 54.6 and 52.7) and precision (50.0 vs. 42.5 and 40.4) metrics also show substantial improvements. 
\subsubsection{Egocentric View}
Similarly, we show comparison results in Table 4 of the main paper. CSTS~\cite{lai2024listen} is the state-of-the art model for egocentric gaze anticipation. However, CSTS~\cite{lai2024listen} is not trained/evaluated on EGTEA Gaze+ as they require audio data. We adapt their backbone model, denoted as ``CSTS-Visual~\cite{lai2024listen}" in Table.~4 to gaze anticipation task on EGTEA Gaze+. 
In addition to CSTS-Visual, we adapt another three existing models for gaze anticipation, denoted as GLC$^\dagger$~\cite{GLC}, MViT$^\dagger$~\cite{fan2021multiscale} and I3D-R50$^\dagger$~\cite{feichtenhofer2019slowfast}. These models were retrained to predict gaze heatmaps at 0.5 seconds into the future. We re-train the four models by using the grounf-truth heatmaps at the 0.5 seconds later. However, their performance was substantially lower compared to SAGE. 


\subsection{Extensive Experimental Analysis} \label{supp: section experiments}
\subsubsection{Model Efficiency.} \label{efficiency} Although efficiency is not our focus, SAGE achieves strong recognition and anticipation accuracy while maintaining high computational efficiency, processing a 16-frame input sequence (on EGTEA Gaze+) in just 0.63 s on an NVIDIA RTX A6000. The full SAGE (egocentric) model contains 333.2M parameters, representing only a modest 3\% increase compared to the combined next-best task-specific baselines (GLC, GC-TSM, GLC-adapted, and InAViT), despite jointly addressing all four tasks within a single architecture. In addition, SAGE exhibits comparable inference efficiency, with a slightly faster runtime (by 2\%) than independently executing the four task-specific models.
We also compare the computational complexity for one round of iterative inference of SAGE vs other SOTA methods in terms of GFLOPS (Giga Floating-Point Operations Per Second) in Table~\ref{glfops}. 
As shown, the proposed model achieves higher accuracy while reducing computational complexity by 11.3\% compared to the \textit{combined} next-best task-specific baselines. This demonstrates an improved accuracy–compute tradeoff, delivering better predictive performance at lower computational cost, enabling enhanced suitability for real-world deployment under compute and power constraints.

\begin{table}[]
\centering
\caption{Comparison of computational cost (GFLOPs). SAGE achieves an 11.3\% reduction in total GFLOPs compared to the combined cost of prior SOTA models.}
\label{glfops}
\setlength{\tabcolsep}{4pt}
\begin{tabular}{c|cc}
\toprule
\multirow{2}{*}{Model} & \multicolumn{2}{c}{GFLOPS}          \\ \cline{2-3} 
                       & Individual & Total                  \\ \hline
GLC~\cite{GLC}                    & 72.2       & \multirow{4}{*}{601.5} \\
GC-TSM~\cite{hao2022group}                 & 66.5       &                        \\
GLC~\cite{GLC}                    & 70.8       &                        \\
InAViT~\cite{InAViT}                 & 392        &                       \\ \hline
SAGE (Ours)                   & 533.4      & 533.4\,{\color{ForestGreen}{$\scriptstyle\downarrow 11.3\%$}}                \\
\bottomrule
\end{tabular}
\end{table}

\subsubsection{Sensitivity of SAGE to Gaze Model.}\label{supp: section egtea}
\label{appendix_egtea}
\begin{table*}[]
\small
\centering
\setlength{\tabcolsep}{2pt}
\caption{Sensitivity of SAGE to gaze modules on the EGTEA Gaze+ dataset}
\begin{tabular}{c|c|ccc|c|ccc|c}
\toprule
\multirow{2}{*}{Model} & \multirow{2}{*}{Gaze Model} & \multicolumn{3}{c|}{\begin{tabular}[c]{@{}c@{}}Gaze \\ Detection\end{tabular}} & \begin{tabular}[c]{@{}c@{}}Action \\ Recogn.\end{tabular} & \multicolumn{3}{c|}{\begin{tabular}[c]{@{}c@{}}Gaze \\ Anticipation\end{tabular}} & \begin{tabular}[c]{@{}c@{}}Action \\ Anticipation\end{tabular} \\ \cline{3-10} 
                        &                             & F1                 & Rec                         & Prec                        & Top-1 Acc                                                 & F1                        & Rec                       & Prec                      & Mean-cls Acc                                                   \\ \hline
\multirow{3}{*}{SAGE}   & I3D-R50\text{~\cite{feichtenhofer2019slowfast}}                     & 42.2               & 58.5                        & 33.2                        & 64.6                                                      & 34.7                      & 47.4                      & 30.2                      & 57.8                                                           \\
                        & MViT\text{~\cite{fan2021multiscale}}                        & 44.5               & 58.8                        & 35.5                        & 64.6                                                      & \textbf{38.8}                      & 54.2                      & 32.5                      & 58.0                                                           \\
                        & GLC\text{~\cite{GLC}}                         & \textbf{46.8}               & \textbf{62.1}               & \textbf{36.8}               & \textbf{65.0}                                                      & 37.0                      & \textbf{54.9}                      & \textbf{32.7}                      & \textbf{58.4}  \\
\bottomrule
\end{tabular}
\label{table:ablation_gazesenitivity}
\end{table*}

In the main paper, we compare the performance of SAGE with state-of-the-art models across different task domains. In the exocentric view, the effectiveness of gaze is more apparent, as the gaze model can explicitly identify the visual target of the person within the scene. Intuitively, the gaze-based attention closely aligns with the object being interacted with, making it a strong prior for modeling human-object interactions. In Table 1 of the main paper, we show the sensitivity of SAGE performance to the accuracy of gaze model. We incorporate two different gaze backbone model, denoted as SAGE (Gaze Detection Model: VideoAttn) and
SAGE (Gaze Detection Model: Sharingan) in Table 1, into SAGE architecture. The results show that better gaze detection model in SAGE enhances performances for other three tasks.
\\
Similarly, to explore the sensitivity of SAGE to the quality of gaze estimation in egocentric videos, we compare three different gaze backbones in Table~\ref{table:ablation_gazesenitivity}: I3D-R50~\cite{feichtenhofer2019slowfast}, MViT~\cite{fan2021multiscale}, and GLC~\cite{GLC}. We observe that improvements in gaze detection performance consistently lead to better action recognition and anticipation results. For instance, GLC achieves the best gaze detection F1 score (46.8), which corresponds to the highest top-1 accuracy in action recognition (65.0) and action anticipation (58.4). In contrast, models with weaker gaze detection performance, such as I3D-R50 (F1 = 42.2), yield the lowest accuracy in both action recognition and anticipation. These results suggest that precise gaze localization is crucial for enhancing action detection and anticipation in egocentric settings. Intuitively, gaze reflects the subject's focus of interest and serves as an informative prior for modeling human-object interactions. The results in Table~\ref{table:ablation_gazesenitivity} prove that better gaze attention enhances the performance for each task. It validates the effectiveness of our GCSA module. 
\subsubsection{Sensitivity of SAGE to Anticipation Horizon.} \label{supp anticipation horizon}
Following InAViT~\cite{InAViT}, we explore SAGE performance on longer time horizons. In Table~\ref{table:supp longer_horizon}, we train SAGE on different time horizons \{0.5s, 1s, 1.5s, 2s\}. Notably, SAGE consistently outperforms InAViT on all time horizons. 
As the anticipation horizon increases, there are greater uncertainty for future action prediction. The Top-1 accuracy of action anticipation in InAViT decreases from 67.8 to 64.1 when the anticipation time increases to 2.0s. As SAGE is anticipating future actions with multiple intermediate time stamps (i.e., $[y_{t+1},\cdots,y_{t+M}]$), SAGE maintains more stable performance over longer time spans.
\begin{table}[]
\small
\centering
\setlength{\tabcolsep}{2pt}
\caption{Top-1 accuracy for SAGE at longer anticipation horizon on the EGTEA Gaze+ dataset.}
\begin{tabular}{c|cccc}
\toprule
\multirow{2}{*}{Method} & \multicolumn{4}{c}{$\tau_a$}  \\ \cline{2-5} 
                        & 0.5   & 1.0   & 1.5   & 2.0  \\ \hline
InAViT~\cite{InAViT}    & 67.8  & 66.9  & 65.8  & 64.1 \\ \hline
SAGE (Ours)                   & \textbf{68.0}  & \textbf{67.3}  & \textbf{66.2}  & \textbf{64.5}\\
\bottomrule
\end{tabular}
\label{table:supp longer_horizon}
\end{table}

\subsubsection{Extensive Results On Vid-HOI.}
\begin{table}[]
\small
    \centering
    \setlength{\tabcolsep}{1pt}
    \caption{HOI detection in Oracle mode on Vid-HOI dataset~\cite{vid-hoi}. * includes word embedding module.}
    \begin{tabular}{c c c c c c c}
        \toprule
        Method & $\tau_a$ & mAP & Rec & Prec & Acc & F1 \\
        \midrule
        \multirow{1}{*}{STTran~\cite{cong2021spatial}} 
        & - & 28.32 & - & - & - & - \\
        \multirow{1}{*}{ST-Gaze*~\cite{ni2023human}} 
        & - & 38.46 & \textbf{73.62} & 59.16 & \textbf{53.76} & 60.57 \\
        \midrule
        \multirow{1}{*}{ST-Gaze Spatial*~\cite{ni2023human}} 
        & - & 36.29 & 71.03 & 59.38 & 51.72 & 61.24 \\
        \multirow{1}{*}{SAGE-12 (Ours) } 
        & - & \textbf38.08 & 72.11 & 59.90 & 52.12 & \textbf{62.03} \\
        \midrule
        \multirow{4}{*}{SAGE (Ours)} 
        & 1 & \textbf{38.65} & 72.44 & 59.22 & 52.22 & 61.96 \\
        & 3 & 38.18 & 72.21 & \textbf{59.95} & 52.18 & 62.02  \\
        & 5 & 38.13 & 72.01 & 59.92 & 52.18 & 61.88 \\
        \bottomrule
    \end{tabular}
    \label{table:supp-vid}
\end{table}\label{supp: section vidhoi}
In Table~\ref{table:supp-vid}, we benchmark HOI detection performance on the Vid-HOI dataset using ST-Gaze~\cite{ni2023human} and STTran~\cite{cong2021spatial} as baselines. Since Vid-HOI does not provide gaze labels, gaze evaluation is omitted. To specifically evaluate the impact of Gaze-Conditioned Attention in the spatial domain, we introduce a spatial-only variant of ST-Gaze by removing its temporal layer (referred to as ST-Gaze Spatial). Additionally, we include SAGE-12 in the comparison. Notably, SAGE-12 achieves the highest F1 score (62.03), underscoring the importance of integrating gaze-based attention mechanisms into HOI recognition tasks. Furthermore, our full SAGE model attains the best mAP (38.65) and Recall (72.44) scores at the 1-second anticipation setting, and the highest Precision (59.95) at the 3-second setting. 

\subsubsection{SAGE Ablation Study} \label{supp: sage ablation causal}
In the main paper, we show ablation study of model combination on an egocentric dataset, i.e. EGTEA Gaze+ dataset. In this section, we provide a more comprehensive analysis of different component of SAGE.
We first introduce different combination of SAGE modules. 
In Section 4 of the main paper, we decompose the problem of joint estimation of current and future action as:
\begin{equation}
    \begin{aligned}
        &p(\bm{y}_{t:t+M} \mid I_{0:t}) = \int_G p(\bm{y}_{t:t+M}, \bm{g}_{0:t+M} \mid I_{0:t}) \, dG \\
        &= \int_G p(\bm{y}_{t:t+M} \mid \bm{g}_{0:t+M}, I_{0:t}) \, p(\bm{g}_{0:t+M} \mid I_{0:t}) \, dG \\
        &= \int_G p(y_t \mid \bm{g}_{0:t}, I_{0:t}) \, p(\bm{y}_{t+1:t+M} \mid y_t, \bm{g}_{t+1:t+M}, I_{0:t}) \\
        &\quad \cdot \, p(\bm{g}_{0:t} \mid I_{0:t}) \, p(\bm{g}_{t+1:t+M} \mid \bm{g}_{0:t}, I_{0:t}) \, dG
    \end{aligned}
\label{supp: eq1}
\end{equation}
\textbf{Causal Assumption.} In Section~4 of the main paper, we adopt a simplifying causal assumption that future actions $\bm{y}_{t+1:t+M}$ are conditionally independent of past gaze $\bm{g}_{0:t}$ given the predicted future gaze $\bm{g}_{t+1:t+M}$. This leads to the approximation
$$p(\bm{y}_{t+1:t+M} \mid y_t, \bm{g}_{0:t+M}, I_{0:t})
\approx
p(\bm{y}_{t+1:t+M} \mid y_t, \bm{g}_{t+1:t+M}, I_{0:t}).$$
The assumption is introduced to prevent an unnecessarily larger model architecture, as it removes the need to explicitly propagate the full historical gaze sequence into the action anticipation module. Intuitively, once the current action representation and the predicted future gaze trajectory are available, historical gaze observations provide limited additional information for predicting future actions.

To validate this assumption, we evaluate a variant that conditions future action prediction on the full gaze sequence $\bm{g}_{0:t+M}$. As shown in Table~\ref{supp:table:causal}, the variant (denoted as SAGE$^*$) performs slightly worse than the proposed SAGE model in both gaze anticipation and action anticipation, while introducing additional model complexity. Therefore, in our final implementation, the action anticipation module only conditions on future gaze encodings.
\begin{table}[]
\small
\centering
\setlength{\tabcolsep}{1pt}
\caption{Empirical validation of the conditional independence assumption between past gaze and future actions.}
\scalebox{0.95}{
\begin{tabular}{c|ccc|c|ccc|c}
\toprule
\multirow{2}{*}{Models} & \multicolumn{3}{c|}{\begin{tabular}[c]{@{}c@{}}Gaze \\ Detection\end{tabular}} & \begin{tabular}[c]{@{}c@{}}Action \\ Recogn.\end{tabular} & \multicolumn{3}{c|}{\begin{tabular}[c]{@{}c@{}}Gaze \\ Anticipation\end{tabular}} & \begin{tabular}[c]{@{}c@{}}Action \\ Anticipation\end{tabular} \\ \cline{2-9} 
                        & F1                       & Rec                      & Prec                     & Acc                                                         & F1                        & Rec                       & Prec                      & Acc                                                            \\ \hline

SAGE$^*$ ($\bm{y}_{t+1:t+M}|\bm{g}_{0:t+M},y_t$) & 46.6 & 62.0 & 36.5 & 65.0 & 36.3 & 54.2 & 32.3 & 57.7 \\

SAGE  ($\bm{y}_{t+1:t+M}|\bm{g}_{t+1:t+M},y_t$)                  & \textbf{46.8}                     & \textbf{62.1}                     & \textbf{36.8}                     & 65.0                                                       & \textbf{37.0}                      & \textbf{54.9}                      & \textbf{32.7}                      & \textbf{58.4} \\
\bottomrule
\end{tabular}
}
\label{supp:table:causal}
\end{table}

In the main paper, we analyze the distinct effects of the GCSA and GCTP modules by reporting the performance of SAGE-12 and SAGE-34. Beyond these settings, SAGE can also be configured in additional combinations to perform specific tasks with or without the GCSA and GCTP components. Here, we extend the ablation study from the main paper by exploring two additional SAGE configurations.
\begin{itemize}
    \item \textbf{SAGE-123:} The model recognizes current actions conditioned on current gaze and additionally predicts future gaze based on the observed gaze trajectory. However, as formulated in Eq.~\ref{supp: eq1}, the future gaze variables $\bm{g}_{t+1:t+M}$ are redundant for the action recognition task, since we assume that given $\bm{g}_{0:t}$, the current action $y_t$ is conditionally independent of $\bm{g}_{t+1:t+M}$. We include this configuration to investigate the potential correlation between future gaze and future actions.
    \item \textbf{SAGE-124:} The model recognizes both current and future actions conditioned on current gaze, without involving gaze anticipation. The GCTP module is not activated. Given the formulation of Eq~\ref{supp: eq1}, we need to compute future actions conditioning on current action and current gaze features. We random initialize a list of learnable tokens to replace the future gaze encoding in the GCTP module.  
\end{itemize}

\subsubsection{Ablation Study on EGTEA Gaze+.}
\begin{table}[]
\small
\centering
\setlength{\tabcolsep}{1pt}
\caption{SAGE ablations of GCSA and GCTP on EGTEA Gaze+ dataset.}
\scalebox{0.95}{
\begin{tabular}{c|ccc|c|ccc|c}
\toprule
\multirow{2}{*}{Models} & \multicolumn{3}{c|}{\begin{tabular}[c]{@{}c@{}}Gaze \\ Detection\end{tabular}} & \begin{tabular}[c]{@{}c@{}}Action \\ Recogn.\end{tabular} & \multicolumn{3}{c|}{\begin{tabular}[c]{@{}c@{}}Gaze \\ Anticipation\end{tabular}} & \begin{tabular}[c]{@{}c@{}}Action \\ Anticipation\end{tabular} \\ \cline{2-9} 
                        & F1                       & Rec                      & Prec                     & Acc                                                         & F1                        & Rec                       & Prec                      & Acc                                                            \\ \hline
\color[HTML]{C0C0C0} {SAGE-1}                  & \color[HTML]{C0C0C0} 44.8                     & \color[HTML]{C0C0C0} 61.2                     & \color[HTML]{C0C0C0} 35.3                     & -                                                           & -                         & -                         & -                         & -                                                              \\
SAGE-2                  & -                        & -                        & -                        & 63.1                                                        & -                         & -                         & -                         & -                                                              \\
SAGE-12                 & 46.3                     & 61.9                     & 36.4                     & 63.9                                                        &                           &                           &                           &                                                                \\
SAGE-123 & 46.2 & 61.5 & 36.2 & 64.1 & 36.5 & 54.3 & 32.1 & - \\
SAGE-124 & 45.8 & 61.7 & 36.3 & 64.4 & - & -& -& 56.3\\
SAGE                    & \textbf{46.8}                     & \textbf{62.1}                     & \textbf{36.8}                     & \textbf{65.0}                                                        & \textbf{37.0}                      & \textbf{54.9}                      & \textbf{32.7}                      & \textbf{58.4} \\
\bottomrule
\end{tabular}
}
\label{table:ablation-egtea}
\end{table}

\begin{table}[t]
\small
\centering
\setlength{\tabcolsep}{3pt}
\caption{Effectiveness of GCTP on EGTEA Gaze+ across different anticipation horizons $\tau_a$.}
\scalebox{0.9}{
\begin{tabular}{l c|ccc|c|c}
\toprule
\multirow{2}{*}{Models} & \multirow{2}{*}{$\tau_a$} &
\multicolumn{3}{c|}{\begin{tabular}[c]{@{}c@{}}Gaze \\ Detection\end{tabular}} &
\begin{tabular}[c]{@{}c@{}}Action \\ Recogn.\end{tabular} &
\begin{tabular}[c]{@{}c@{}}Action \\ Anticipation\end{tabular} \\
\cline{3-7}
& & F1 & Rec & Prec & Acc & Acc \\
\midrule
\multirow{4}{*}{SAGE-124} 
& 0.5 & 45.8 & 61.7 & 36.3 & 64.4 & 56.3 \\
& 1.0 & 45.6 & 61.5 & 36.2 & 64.1 & 55.7 \\
& 1.5 & 45.9 & 61.8 & 36.5 & 64.1 & 54.6 \\
& 2.0 & 45.6 & 61.2 & 36.3 & 63.6 & 53.2 \\
\midrule
\multirow{4}{*}{SAGE} 
& 0.5 & 46.8 & 62.1 & 36.8 & 65.0 & 58.4 \\
& 1.0 & 46.7 & 62.0 & 36.7 & 64.7 & 57.4 \\
& 1.5 & 46.2 & 61.9 & 36.2 & 64.1 & 56.3 \\
& 2.0 & 46.5 & 62.5 & 36.5 & 63.7 & 56.2 \\
\bottomrule
\end{tabular}
}
\label{table:ablation_time_horizon_egtea}
\end{table}

\begin{table}[t]
\small
\centering
\setlength{\tabcolsep}{3pt}
\caption{SAGE Ablations on EGTEA Gaze+ of how gaze anticipation performance affected by future actions across different anticipation horizons $\tau_a$.}
\scalebox{0.9}{
\begin{tabular}{l c|ccc|c|c c c}
\toprule
\multirow{2}{*}{Models} & \multirow{2}{*}{$\tau_a$} &
\multicolumn{3}{c|}{\begin{tabular}[c]{@{}c@{}}Gaze \\ Detection\end{tabular}} &
\begin{tabular}[c]{@{}c@{}}Action \\ Recogn.\end{tabular} &
\multicolumn{3}{c}{\begin{tabular}[c]{@{}c@{}}Gaze \\ Anticipation\end{tabular}} \\
\cline{3-9}
& & F1 & Rec & Prec & Acc & F1 & Rec & Prec \\
\midrule
\multirow{4}{*}{SAGE-123} 
& 0.5 & 46.2 & 61.5 & 36.2 & 64.1 & 36.5 & 54.3 & 32.1 \\
& 1.0 & 46.1 & 61.4 & 36.2 & 63.8 & 35.7 & 53.7 & 31.5 \\
& 1.5 & 45.6 & 61.5 & 36.2 & 63.2 & 35.1 & 52.8 & 31.0 \\
& 2.0 & 45.6 & 61.5 & 36.2 & 62.6 & 34.4 & 51.9 & 30.2 \\
\midrule
\multirow{4}{*}{SAGE} 
& 0.5 & 46.8 & 62.1 & 36.8 & 65.0 & 37.0 & 54.9 & 32.7 \\
& 1.0 & 46.7 & 62.0 & 36.7 & 64.7 & 36.1 & 54.0 & 32.0  \\
& 1.5 & 46.2 & 61.9 & 36.2 & 64.1 & 35.5 & 53.6 & 31.8 \\
& 2.0 & 46.5 & 62.5 & 36.5 & 63.7 & 34.9 & 52.2 & 30.6\\
\bottomrule
\end{tabular}
}
\label{table:ablation_SAGE123_egtea}
\end{table}

In Table~\ref{table:ablation-egtea}, we conduct ablations on an egocentric dataset, i.e. EGTEA Gaze+ dataset.\\
\noindent\textbf{GCSA effectiveness.}
Incorporating the GCSA module significantly improves both gaze detection and action recognition in Table~\ref{table:ablation-egtea}. Compared to SAGE-2, which performs action recognition without gaze conditioning, SAGE-12 achieves higher accuracy (63.9\% vs. 63.1\%) and better gaze detection (46.3 F1 vs. 44.8). 
This demonstrates that gaze attention provides discriminative spatial priors for action understanding and further stabilizes gaze localization. 
\\
\textbf{GCTP effectiveness.}
In Table~\ref{table:ablation-egtea}, we compare SAGE-124 with the full SAGE model (with GCTP). While SAGE consistently outperforms SAGE-124 across all tasks, the improvement is particularly notable for action anticipation. This shows that the GCTP module effectively captures predictive temporal dependencies between future gaze and future actions.
In addition, we also study how future gaze contributes to action anticipation on longer time horizons. We compare the results of SAGE-124 with the full SAGE model on additional $\tau_a$ (1.0s, 1.5s and 2.0s) as shown in Table~\ref{table:ablation_time_horizon_egtea}. 
As $\tau_a$ increases, action anticipation accuracy decreases notably for both SAGE-124 and SAGE, reflecting the growing uncertainty of long-term prediction. 
Nevertheless, using future gaze for action anticipation in SAGE consistently achieves better performances than SAGE-124 across all horizons, validating the contribution of GCTP module.

We also analyze how gaze anticipation performance is affected by the presence of the action anticipation task across different anticipation horizons $\tau_a$ in Table~\ref{table:ablation_SAGE123_egtea}. In SAGE-123, no action anticipation is performed, and the gaze anticipation module learns independently of future action supervision. 
In contrast, SAGE jointly optimizes both gaze and action anticipation, where the action anticipation module is conditioned on the future gaze encoding, allowing the action loss to propagate gradients back through the gaze anticipation pathway. 
This joint optimization through SAGE provides an additional supervisory signal that slightly improves gaze anticipation quality over SAGE-123 (e.g., F1 of 37.0 vs.\ 36.5 at $\tau_a = 0.5s$). 
Additionally, as $\tau_a$ increases, the uncertainty in long-term temporal prediction of future gaze grows, leading to degraded gaze anticipation performance for both SAGE and SAGE-123.
\subsubsection{Ablation Study on Exo-Cook}
\begin{table}[]
\small
\centering
\setlength{\tabcolsep}{1.3pt}
\caption{SAGE ablations of GCSA and GCTP on Exo-Cook Dataset.}
\begin{tabular}{c|ccc|c|ccc|c}
\toprule
\multirow{2}{*}{Models} & \multicolumn{3}{c|}{\begin{tabular}[c]{@{}c@{}}Gaze \\ Detection\end{tabular}} & \begin{tabular}[c]{@{}c@{}}HOI \\ Detn.\end{tabular} & \multicolumn{3}{c|}{\begin{tabular}[c]{@{}c@{}}Gaze \\ Anticipation\end{tabular}} & \begin{tabular}[c]{@{}c@{}}HOI \\ Anticipation\end{tabular} \\ \cline{2-9} 
                        & F1                       & Rec                      & Prec                     & Acc                                                         & F1                        & Rec                       & Prec                      & Acc                                                            \\ \hline
SAGE-1                  & 56.6                     & 72.8                     & 55.3                     & -                                                           & -                         & -                         & -                         & -                                                              \\
SAGE-2                  & -                        & -                        & -                        & 58.9                                                        & -                         & -                         & -                         & -                                                              \\
SAGE-12                 & 57.5                     & \textbf{73.6}                     & \textbf{56.0}                     & 59.9                                                        &                           &                           &                           &                                                                \\
SAGE-123 & 57.5 & 73.2 & 55.4 & 60.2 & 48.8 & 62.4 & 49.8 & - \\
SAGE-124 & 57.5 & 73.2 & 55.6 & 59.9 & - & -& -& 52.5 \\
SAGE                    & \textbf{57.7}                     & 73.4                     & 55.5                     & \textbf{60.2}                                                        & \textbf{49.2}                      & \textbf{62.8}                      & \textbf{50.0}                      & \textbf{54.8} \\
\bottomrule
\end{tabular}
\label{table:ablation_Exocook}
\end{table}


In Table~\ref{table:ablation_Exocook}, we conduct similar ablations on an exocentric dataset, i.e. Exo-Cook dataset. 
We use Sharingan~\cite{tafasca2024sharingan} as the gaze model (SAGE-1).

\noindent\textbf{GCSA effectiveness.} SAGE-1 and SAGE-2 perform gaze detection and HOI detection (oracle mode) independently and we observe that combining them into the joint model SAGE-12 through GCSA leads to consistent improvements in both the tasks, which effectively integrates gaze into spatial attention, allowing the model to better capture human-object interactions. 

\noindent\textbf{GCTP effectiveness.} We observe that SAGE outperforms SAGE-124 in action anticipation, indicating that the GCTP module effectively captures predictive temporal dependencies between future gaze and future actions. We also find that SAGE achieves slightly better gaze anticipation results than SAGE-123, owing to the joint optimization of both future gaze and future action—consistent with the trend observed on the EGTEA Gaze+ dataset (Table~\ref{table:ablation_SAGE123_egtea}).
It should be noted that since the optimal SAGE model is selected based on best HOI detection performance, the checkpoint yielding the best gaze detection result may differ. As shown in Table 5, the F1 and Recall of SAGE are slightly lower than those of SAGE-12, which also suggests that gaze tends to be noisier and less directly linked to actions in exocentric videos. Overall, the progressive improvements from SAGE-1/2 to SAGE-12 and finally to the full SAGE model demonstrates that both GCSA and GCTP are essential for fully leveraging the bidirectional relationship between gaze and HOIs in both current and future frames.


\begin{figure*}[ht!]
    \centering
    \includegraphics[width=\textwidth, height=5.1in]{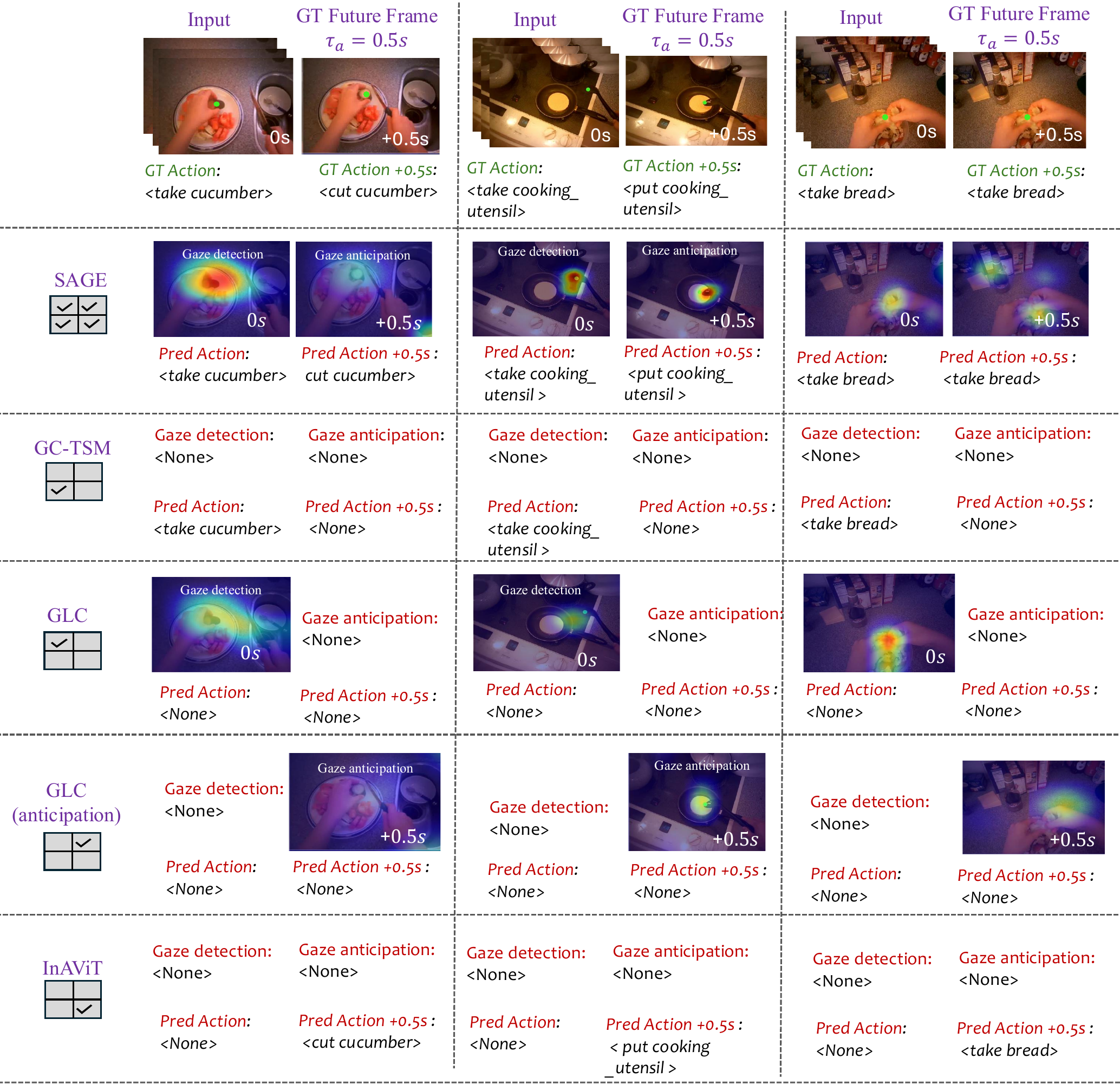}
    \caption{Visualization of SAGE results on the EGTEA Gaze+ dataset, illustrating action and gaze estimation for both current and future frames. The leftmost column displays checkbox representation corresponding to each method, indicating the output settings available. Our method (SAGE) supports all four prediction settings: gaze detection, action recognition, gaze anticipation, and action anticipation.}
    \label{fig:supp_vis_EGTEA}

\end{figure*}

\begin{figure*}[ht!]
    \centering
    \includegraphics[width=\textwidth, height=4.8in]{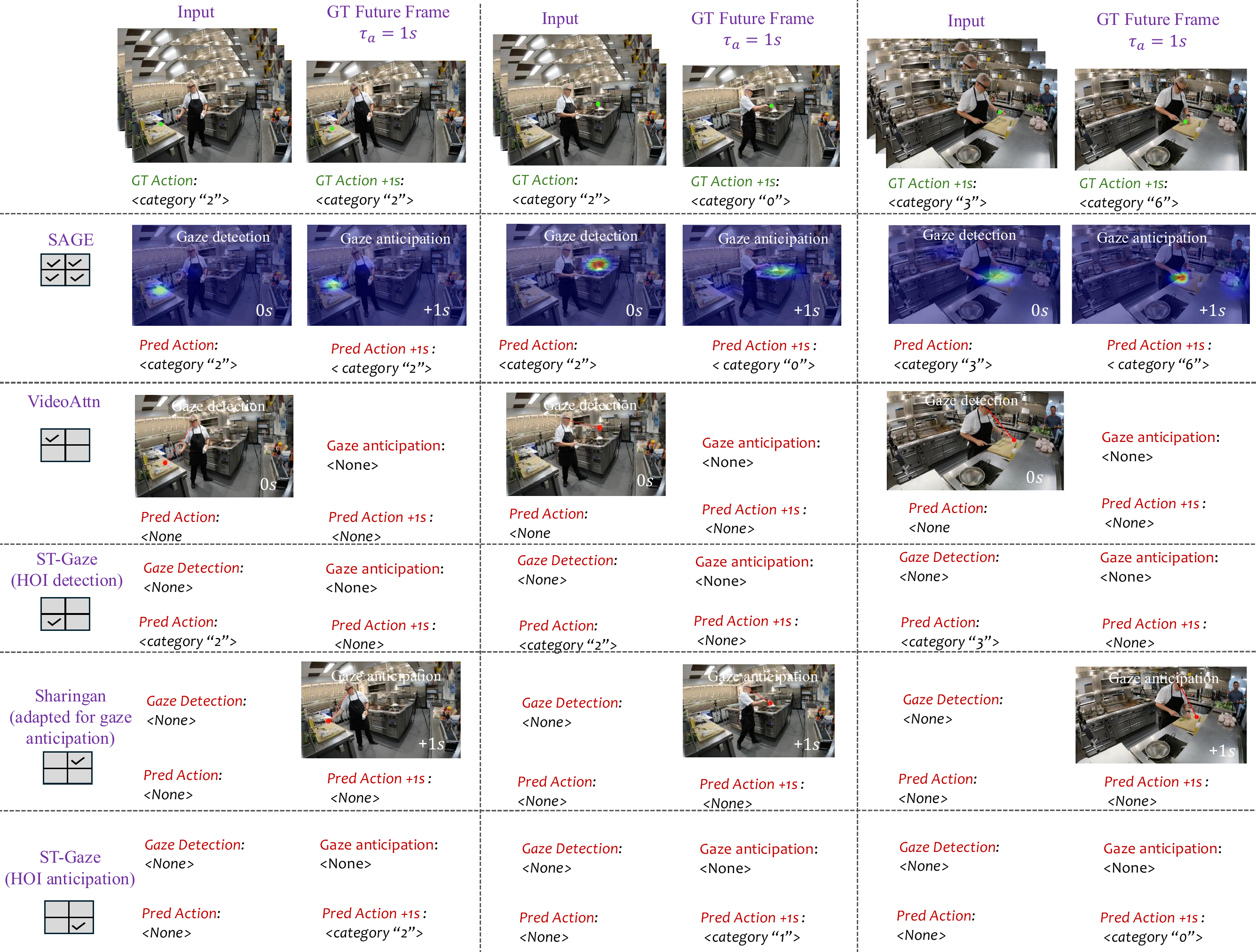}
    \caption{Visualization of SAGE results on Exo-Cook, including action and gaze results for current and future frames. The leftmost column displays checkbox representation corresponding to each method, indicating the output settings available. Our method (SAGE) supports all four prediction settings: gaze detection, action recognition, gaze anticipation, and action anticipation.}
    \label{fig:supp_vis_exocook}
\end{figure*}
\begin{figure}[ht!]
\centering

\begin{subfigure}{0.98\linewidth}
    \centering
    \includegraphics[width=\linewidth]{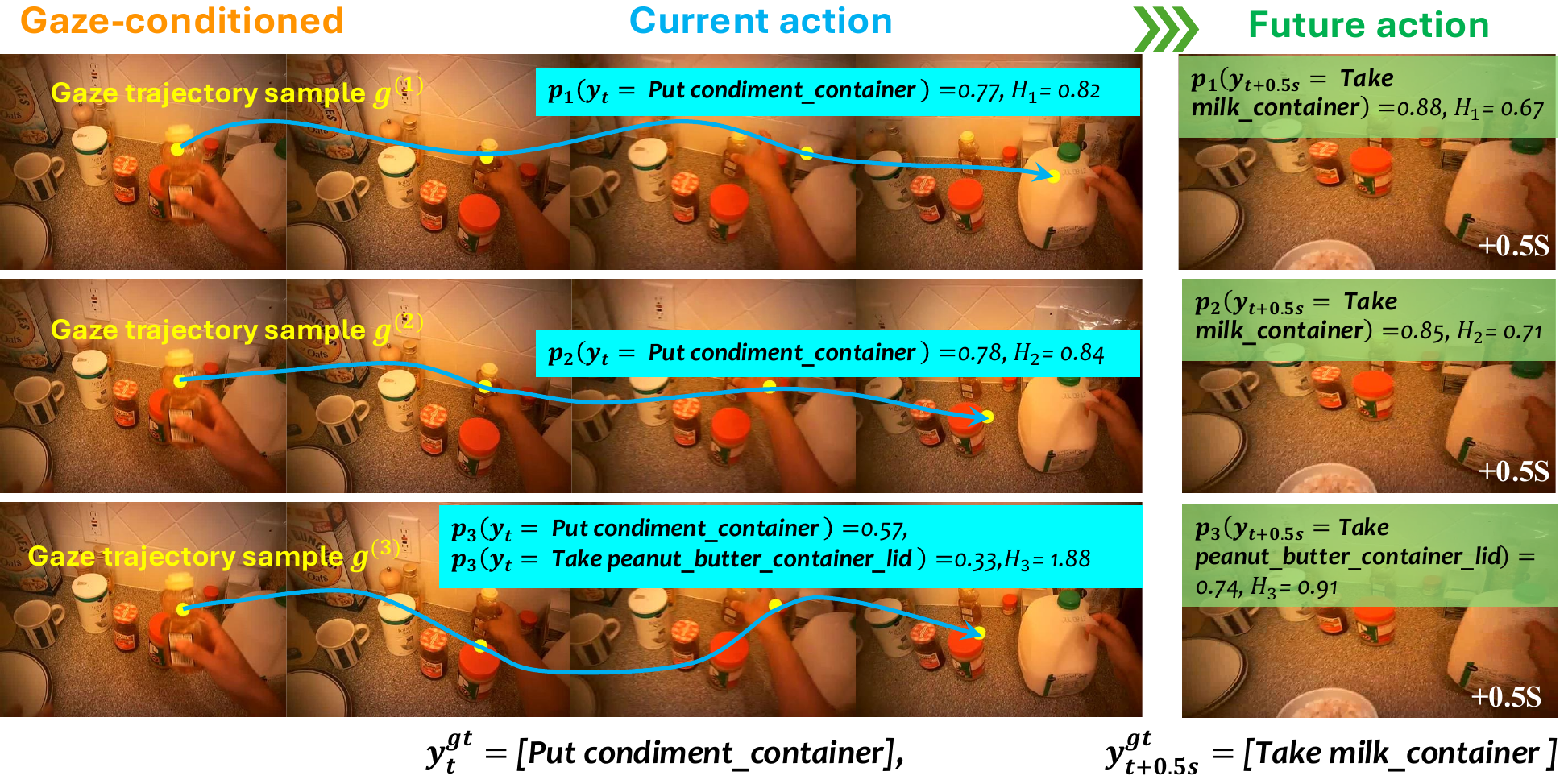}
    \caption{Gaze-conditional action uncertainty example from EGTEA Gaze+.}
\end{subfigure}

\vspace{4mm}

\begin{subfigure}{0.98\linewidth}
    \centering
    \includegraphics[width=\linewidth]{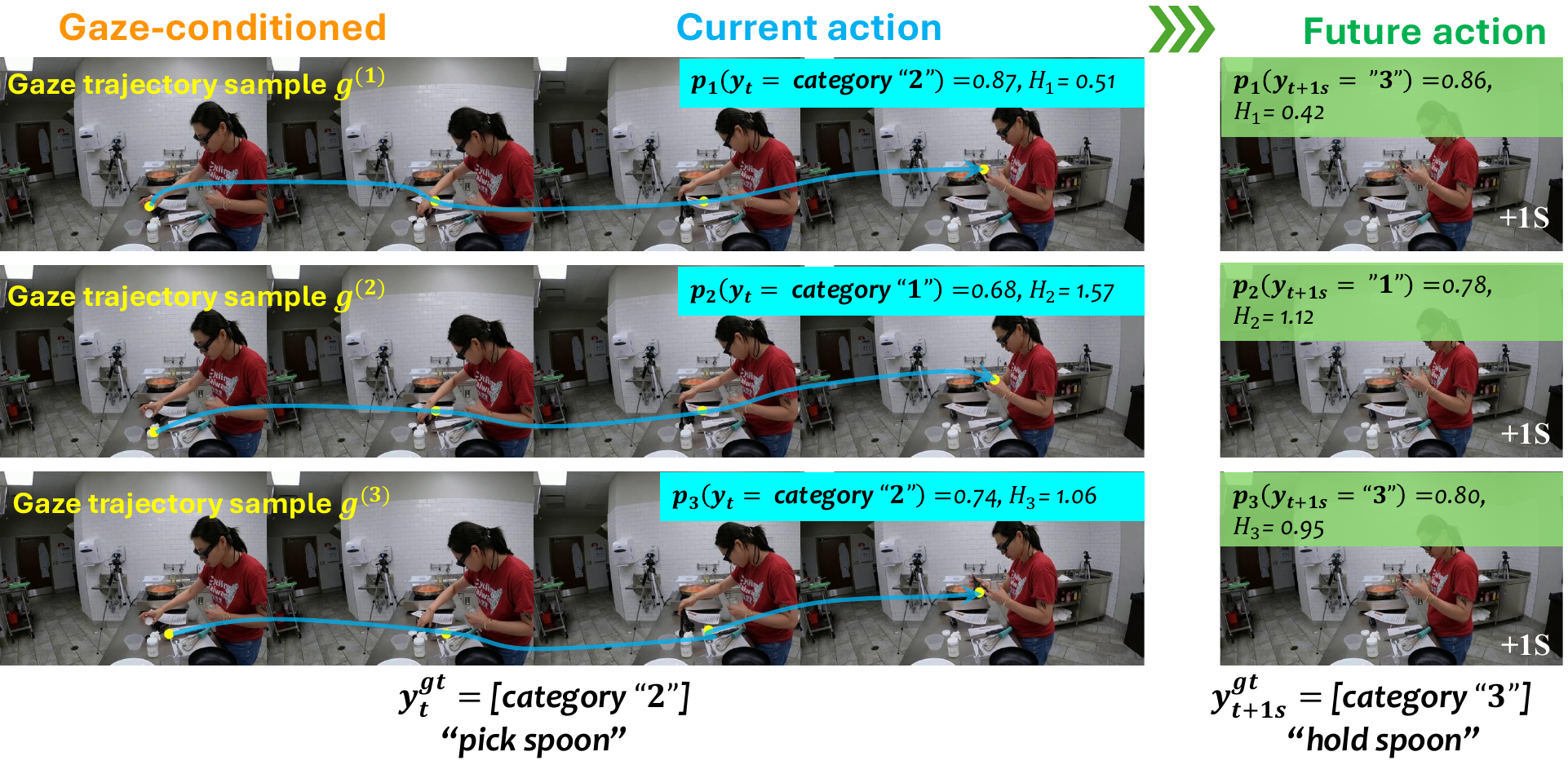}
    \caption{Gaze-conditional action uncertainty example from Exo-Cook.}
\end{subfigure}
\caption{Qualitative visualization of gaze-conditioned action sensitivity. For each clip, three sampled gaze trajectories are used to predict the current action $y_t$ and future action $y_{t+M}$, with corresponding probabilities and entropy $H$. (a) On EGTEA Gaze+, action predictions vary noticeably with gaze, with certain gaze samples increasing uncertainty and causing confusion between similar actions. (b) On Exo-Cook, different gaze samples lead to distinct predictions; misleading intermediate fixations can result in incorrect current and future action predictions.}
\label{supp:fig:action uncertainty}
\end{figure}
\subsection{Qualitative Results of SAGE}\label{supp: section vis}
Figure~\ref{fig:supp_vis_EGTEA} provides qualitative comparisons of our proposed model \textbf{SAGE} with four competitive baselines—GC-TSM~\cite{hao2022group}, GLC~\cite{GLC}, GLC (anticipation), and InAViT~\cite{InAViT}—on the EGTEA Gaze+ dataset. The evaluation covers four tasks: gaze detection, gaze anticipation, action recognition, and action anticipation, with an anticipation horizon of $\tau_a = 0.5$ seconds. Each column shows the last frame (at $0s$) in the input sequence and its corresponding ground-truth future frame at $+0.5s$. The left example illustrates a two-step activity transition from \textit{taking a cucumber} to \textit{cutting the cucumber}. The middle example captures the action progression from \textit{taking a cooking utensil} to \textit{putting it down}, while the right example depicts a temporally stable action—\textit{taking bread}.
SAGE can performs all four tasks. The gaze heatmap predicted from SAGE for current and future frame is well-aligned with the interaction object in the scene. In the meantime, SAGE can predict current and future actions. GC-TSM does not model gaze and only predicts the current action, failing to anticipate future actions. GLC produces accurate gaze detection at $0s$, while GLC (anticipation) provides plausible future gaze at $+0.5s$; however, neither model completes all four tasks jointly. Notably, both variants fail in action anticipation. InAViT does not model gaze and can predict future actions. 

In Figure~\ref{fig:supp_vis_exocook}, we compare SAGE with four state-of-the-art models in different task domain—VideoAttn~\cite{chong2020detecting}, ST-Gaze~\cite{ni2023human}, and Sharingan~\cite{tafasca2024sharingan}—using identical input video sequences. Notably, ST-Gaze provides two separate pipelines for HOI detection and HOI anticipation. While these baseline models are typically designed for individual tasks such as gaze detection or HOI classification, SAGE is the only model can jointly performing all four tasks. The figure showcases three representative examples from the Exo-Cook dataset, displaying the last frame at the current time ($0s$) and a future frame at the anticipation horizon $\tau_a = 1s$. The leftmost example illustrates a stable action sequence, while the middle and right examples involve action transitions. Since actions are labeled with categorical indices, we compare category predictions across models. As discussed in the main paper, no existing model is specifically designed for exocentric gaze anticipation; hence, we adapt Sharingan~\cite{tafasca2024sharingan} for this task. Sharingan is the SOTA model for exocentric gaze target detection in images. However, the adapted Sharingan fails to accurately predict future gaze heatmaps across all three examples. Meanwhile, ST-Gaze can perform either HOI detection or anticipation, but not both simultaneously, and it fails to predict the correct HOI category in the rightmost example. These results highlight the advantage of SAGE in jointly modeling spatial and temporal dynamics for comprehensive human activity understanding.

\subsubsection{Action Uncertainty from Gaze} \label{supp action uncertainty}
We visualize how variations in gaze affect both current action recognition and future action anticipation. For each input clip, we draw $K=3$ gaze samples from the predicted gaze distribution and feed them into the current-action and future-action modules. We then compare the resulting action probabilities
$p(y_t \mid g_{0:t}^{k}, I_{0:t})$
and
$p(y_{t+M} \mid y_t, g_{t+1:t+M}^{k}, I_{0:t})$
across samples.

Figure~\ref{supp:fig:action uncertainty} illustrates how variations in gaze influence both current action recognition and future action anticipation. For each clip, we sample multiple gaze trajectories from the predicted gaze distribution and evaluate the corresponding action probabilities.

In Fig.~\ref{supp:fig:action uncertainty} (a), the predicted actions are sensitive to gaze variations.  Different gaze samples produce different probability distributions and entropy values. In particular, when the sampled gaze trajectory falls on the peanut butter container (in sample $g^{(3)}$), the model becomes uncertain and may confuse between \emph{Take-peanut-butter-container-lid} and \emph{Take-milk-container} for the future action. This example shows that gaze provides an important cue for disambiguating object-level interactions.
Figure~\ref{supp:fig:action uncertainty} (b) demonstrates action-gaze dependency in the exocentric setting. When the sampled gaze trajectory temporarily shifts to the recipe sheet (sample $g^{(2)}$), the model predicts an incorrect action category for both the current and future actions. This indicates that misleading gaze observations can propagate through the interaction reasoning process and affect action inference.
The examples demonstrate that SAGE captures how gaze variations influence action predictions and provides an interpretable estimate of gaze-conditioned uncertainty for both action recognition and anticipation.


\bibliographystyle{splncs04}
\bibliography{main}
\end{document}